\documentclass[pdflatex,sn-mathphys-num]{sn-jnl}

\usepackage{graphicx}%
\usepackage{multirow}%
\usepackage{amsmath,amssymb,amsfonts}%
\usepackage{amsthm}%
\usepackage{mathrsfs}%
\usepackage[title]{appendix}%
\usepackage{xcolor}%
\usepackage{textcomp}%
\usepackage{manyfoot}%
\usepackage{booktabs}%
\usepackage{array}%
\usepackage{tabularx}%
\usepackage{algorithm}%
\usepackage{algorithmicx}%
\usepackage{algpseudocode}%
\usepackage{listings}%
\usepackage{tikz}
\usetikzlibrary{arrows.meta, positioning, fit, backgrounds}
\usepackage{float}

\theoremstyle{thmstyleone}%
\theoremstyle{thmstyletwo}%

\theoremstyle{thmstylethree}%

\begin{document}

\title[Article Title]{How much of an LLM-generated clinical corpus is actually new?
A production-scale measurement of content redundancy for provenance classification}

\author*[1,2]{\fnm{Ali H.} \sur{Lazem}}\email{lhl23prg@bangor.ac.uk}
\author[1]{\fnm{William} \sur{Teahan}}\email{w.j.teahan@bangor.ac.uk}

\affil*[1]{\orgdiv{School of Computer Science and Engineering},
\orgname{Bangor University}, \orgaddress{\city{Bangor},
\postcode{LL57 2DG}, \state{Gwynedd}, \country{United Kingdom}}}
\affil[2]{\orgname{University of Thi-Qar}, \orgaddress{\city{Nasiriyah},
\postcode{64001}, \country{Iraq}}}


\abstract{Clinical machine learning increasingly relies on training corpora
generated by large language models (LLMs) rather than annotated by
clinicians, and such corpora are described and reused largely on the basis
of their reported scale. We test whether volume reflects information
content. Analysing the complete output of a multi-agent clinical extraction
pipeline applied to 167{,}034 patient narratives, 2.51 billion generated
tokens across the ten text-bearing channels of an eleven-channel pipeline,
we introduce Provenance-based Redundancy Decomposition, a token-level
classification of the entire output by source. Only 10.9\% of the output is trainable-unique content while 79.4\% is redundant; the redundant content alone is 19.1 times the original clinical text, and raw token count overstates information content by roughly ninefold. The redundancy arises through two distinct mechanisms, verbatim copying of source context into per-item fields, and duplication of generated text across records, of which only the former is losslessly removable. An independent, model-free analysis based on lossless compression confirms the redundancy: the full corpus compresses several times more than its unique-content
subset (2.7--4.7$\times$ across four compressors), and the two mechanisms are
recovered without reference to the provenance labels. One pipeline channel carries almost no redundancy, showing that the level of
redundancy depends on how each channel is structured rather than being a fixed property of LLM extraction. Because
uncorrected redundancy up-weights the longer, more complex presentations
that generate the most items, it skews the token-level training distribution of
the corpus, a distributional property we measure directly. In a controlled
downstream test, de-duplicating the corpus before adaptation improved a clinical
encoder on external disease-recognition benchmarks at equal token budget,
robustly across adaptation depths and replicated on a second benchmark,
confirming that the redundancy carries a measurable cost beyond storage. The
improvement was a consistent overall gain; a rare-disease-specific benefit we
had hypothesised was tested and not reliably observed. We recommend reporting unique-content size, distinguishing the two mechanisms, and de-duplicating before training. The classification tool is released openly.}




\maketitle

\section{Introduction}\label{sec1}

Clinical machine learning increasingly relies on training data that no
clinician annotated. Large language models now generate clinical corpora
at scales unattainable by manual annotation: extracting entities,
relations, and question answer pairs from hundreds of thousands of
patient narratives, producing structured datasets that are then used to
train downstream models, benchmark clinical systems, and curate
resources for reuse \citep{lee2020biobert, gu2021pubmedbert,
brown2020gpt3}. As this practice spreads, the corpora these pipelines
produce are described, shared, and consumed largely on the basis of
their reported scale, the number of tokens or extracted items they
contain.

Reported scale, however, assumes that volume reflects content. Work on data
curation has shown that the \emph{quality} and \emph{distribution} of training
data can matter more than raw quantity \citep{zhou2023lima, wettig2024qurating},
and that large corpora harbour substantial duplication that degrades models
trained on them \citep{lee2022deduplicating}. A synthetic corpus generated by a
model that applies consistent extraction rules to repetitive clinical text is
especially exposed: the same patient narrative may be referenced by many
extracted items, and the same model output may recur across patients, so the
apparent size of the corpus can substantially exceed the distinct information it
carries. This inflation affects how a corpus's scale and training cost are
reported and, for clinical use most consequentially, how its distribution is
read. To our knowledge, the information content of multi-task clinical LLM
output has not been measured against its raw size at production scale.

We address this gap directly. Analysing the output of a multi-agent clinical
extraction pipeline applied to 167{,}034 patient narratives, a corpus of 2.51
billion generated tokens across ten text-bearing channels, we classify every
output token by provenance and quantify how much of the corpus is genuinely
informative versus redundant. We find that only a small fraction is unique
content, that the redundancy arises through two structurally distinct
mechanisms, and that one task channel in the same pipeline carries almost no
redundancy, showing that the amount of redundancy depends on how a channel is
structured rather than being an inherent property of LLM-driven extraction.

This work makes three contributions. First, we provide the first
production-scale measurement of the gap between the apparent and actual
information content of a multi-task clinical LLM corpus, establishing
that raw token count can overstate content by roughly ninefold. Second, we identify and separately quantify two redundancy
mechanisms, verbatim copying of source context into per-item fields,
and duplication of generated content across records, that call for
different mitigations, only one of which is losslessly removable. Third, we translate these findings into actionable practice. We propose a compact reporting standard for synthetic clinical corpora, the Trainable-Unique Ratio and Context-Copy Ratio, and
release an open tool implementing Provenance-based Redundancy
Decomposition (PRD), the token-classification method introduced here, that computes them for any multi-task extraction pipeline, so that information content can be measured rather than assumed. We accompany these with concrete recommendations: reporting unique-content size
alongside raw size, de-duplicating before training, and adopting provenance classification as a routine quality-control step. Finally, in a controlled downstream experiment we show that de-duplicating the
corpus before encoder adaptation improves disease recognition on an external clinical benchmark at equal token budget, confirming that the redundancy carries a measurable training cost and not merely a storage one.

The finding has direct implications for the reproducibility and fairness of
clinical AI. A field that increasingly trains on synthetic corpora needs those
corpora described in terms that reflect their content, and the distributions
they encode understood rather than silently inherited, prerequisites that
measuring actual information content, rather than apparent size, directly
serves.

\section{Results}\label{sec2}

\subsection{Overall information content of the generated corpus}
We processed the complete output of the multi-task extraction pipeline
over all 167{,}034 PMC-Patients narratives \citep{zhao2023large}. The
pipeline emits structured outputs across eleven task channels, of which
ten are text-bearing and one (visualisation-payload assembly) emits only
structured graph elements (nodes, edges, colours, and levels) together
with entity labels that re-reference NER strings; the visualisation
channel carries no newly generated free text and is outside the scope of
a token-provenance analysis. Across the ten text-bearing channels, the
pipeline emitted 2.51 billion tokens of structured output. We classified
every output token by provenance into five categories: \emph{unique source text} (patient narratives, de-duplicated), \emph{unique generated content} (model-produced text appearing once), \emph{duplicated generated content} (model-produced text repeated across records), and \emph{copied context} (source narrative reproduced verbatim inside per-item context fields), with a residual \emph{scaffold} category for identifiers and enumerated metadata.

\begin{figure}[t]
\centering
\includegraphics[width=0.95\linewidth]{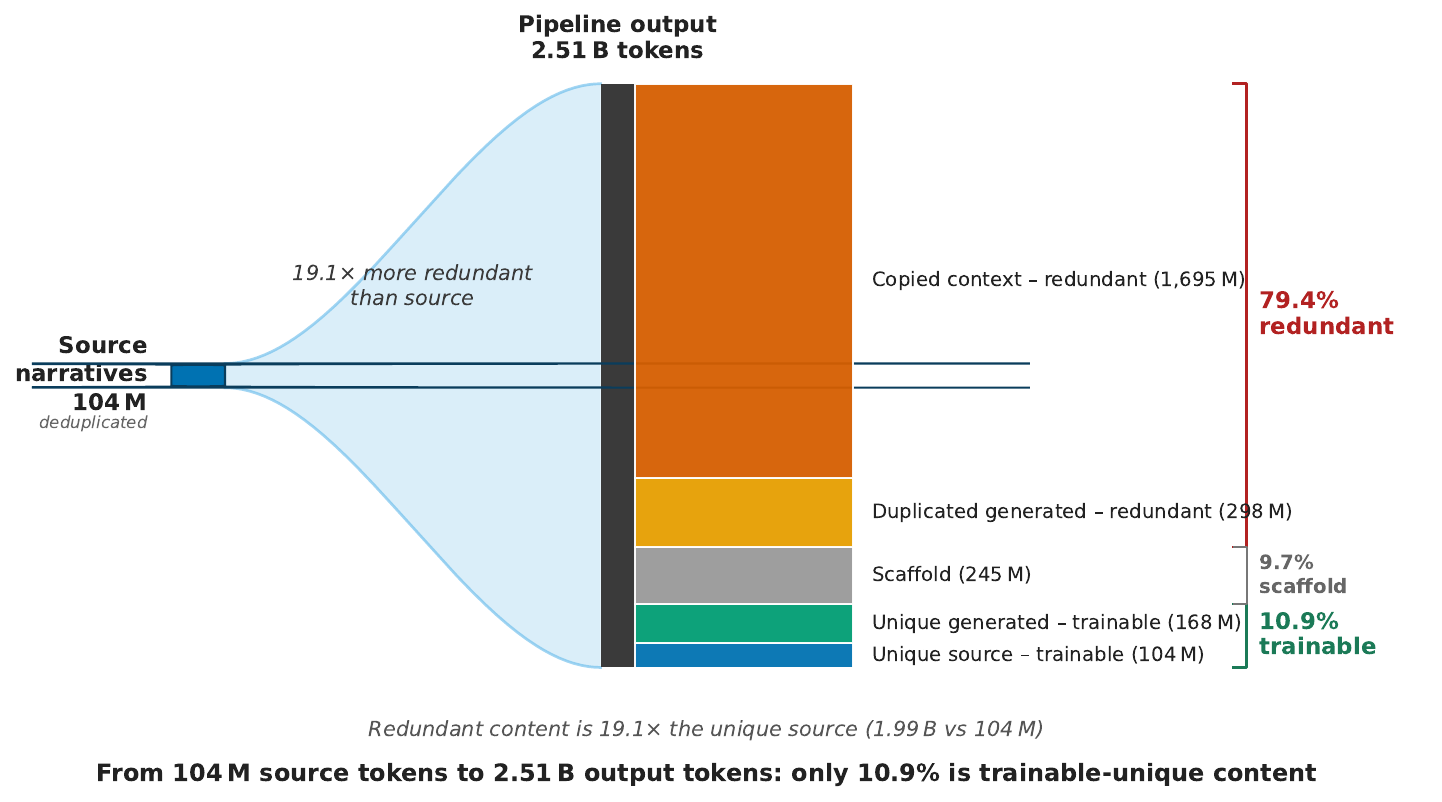}
\caption{\textbf{From source to output.} 104~M unique source tokens
expand into 2.51~B output tokens across ten text-bearing channels, of
which only 10.9\% is trainable-unique content; 79.4\% is redundant
(copied context plus duplicated generation) and 9.7\% is structural
scaffold. The redundant content alone is 19.1 times the unique source.}

\label{fig:global}
\end{figure}

The composition is shown in Figure~\ref{fig:global}. Of the 2.51 billion
output tokens, only 272.6 million (10.9\%) constitute trainable-unique
content, the union of unique source text (104.2~M) and unique generated
content (168.5~M). 1.99 billion tokens (79.4\%) are
redundant: 1.70 billion (67.5\%) are verbatim copies of source narratives
embedded in item-level context fields, and 298.3 million (11.9\%) are
duplicated generated strings; the remaining 244.6 million (9.7\%) are
structural scaffold. The redundant content alone is 19.1 times the
104.2 million tokens of unique source material; the full output is
24.1 times the source, of which the overwhelming majority is redundant
rather than informative.

This result quantifies a gap between the apparent and actual scale of an
LLM-generated clinical corpus. Reported by raw token count, the corpus
appears to contain 2.51 billion tokens of clinical extraction data;
measured by unique content, it contains just 272.6 million. A practitioner
sizing storage, estimating training cost, or describing the corpus in a
data statement using the raw figure would overstate the information content
of the corpus by approximately ninefold (the ratio of the 2.51-billion-token
raw output to the 272.6-million-token trainable-unique content). The larger
expansion factors reported below are stated against a different and smaller
denominator, the 104.2-million-token unique \emph{source} text: the full output
is 24.1$\times$ that source and the redundant content alone is 19.1$\times$ it.
To keep these distinct, we use three denominators throughout, and never
interchange them: the ninefold figure ($2.51\text{B}/272.6\text{M}$) is raw
output over trainable-unique content; the 24.1$\times$ and 19.1$\times$ figures
are full output and redundant content over unique source text; and the
per-patient replication factor reported in
Section~\ref{subsec:results-mechanisms} counts verbatim narrative copies within
a single patient's items.

\subsection{Two distinct redundancy mechanisms}
\label{subsec:results-mechanisms}
\begin{figure}[t]
\centering
\includegraphics[width=0.95\linewidth]{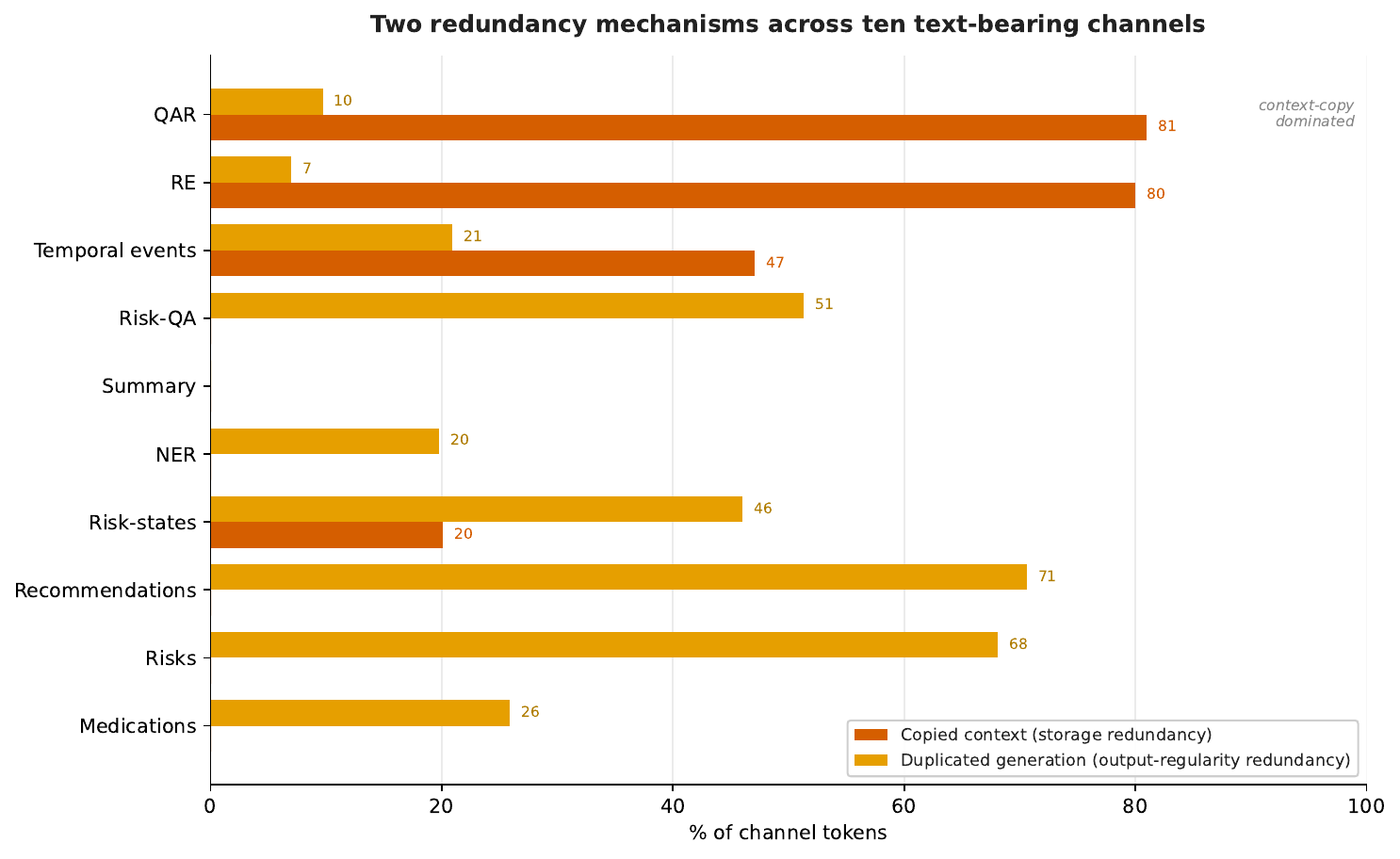}
\caption{\textbf{Two redundancy mechanisms across the ten text-bearing
channels.} Per channel, the percentage of tokens that are copied context
(storage redundancy) versus duplicated generation (output-regularity
redundancy). QAR and RE are context-copy dominated; risk-QA,
recommendations, risks, risk-states, medications, and NER are
generation-duplication dominated; the summary channel is essentially
clean.}

\label{fig:pertask}
\end{figure}

\begin{figure}[t]
\centering
\includegraphics[width=0.95\linewidth]{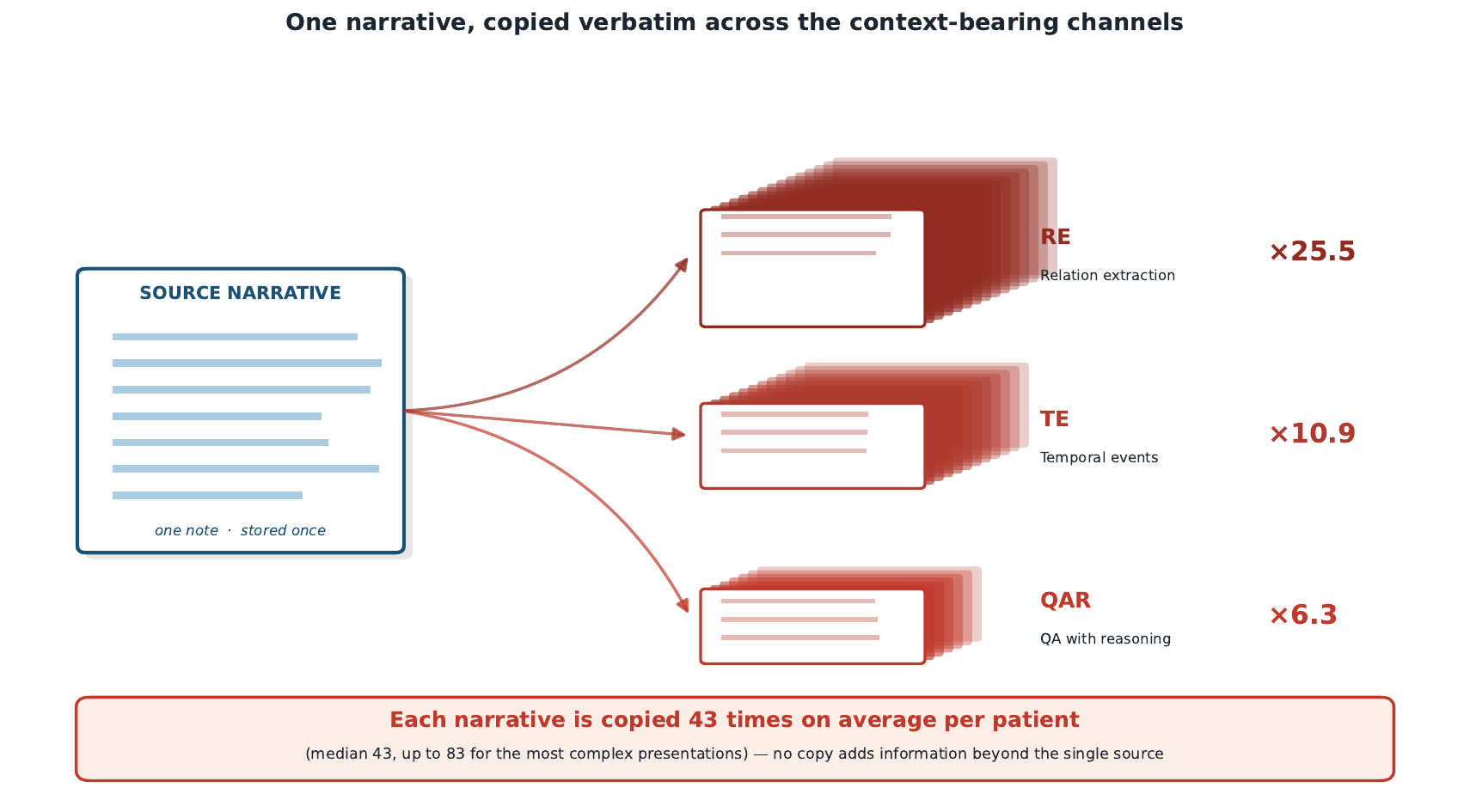}
\caption{\textbf{One narrative, copied verbatim across the context-bearing
channels.} Each patient's source narrative is stored once but reproduced in
full inside the context field of every item the context-bearing channels
emit. The mean number of verbatim copies per patient is 25.5 for relation
extraction, 10.9 for temporal-event extraction, and 6.3 for QAR (corpus
sample, $n = 2{,}000$); the remaining channels (NER, medication, risk-QA,
summary) attach no context and are omitted. Summed across channels, a
single narrative is reproduced on average 42.8 times per patient (median
43, up to 83 for the most complex presentations), with no copy adding
information beyond the single source. Multiplied across 167{,}034 patients,
this per-patient replication is the origin of the corpus-level context-copy
redundancy.}
\label{fig:schematic}
\end{figure}

A single record makes the dominant mechanism concrete: in one representative
QAR item, two verbatim copies of the patient narrative account for 72\% of the
stored text while the extracted fact itself is 0.3\%, shown in detail in
Supplementary Figure~\ref{fig:worked-example}.
Decomposing redundancy by task
(Figure~\ref{fig:pertask}) reveals that this aggregate figure conceals two
structurally different mechanisms, which we term \emph{context-copy redundancy}
and \emph{generation-duplication redundancy}. A worked example of each, drawn
from a single patient record, is given in Supplementary
Table~\ref{tab:mechanisms}
(the same record used in Figure~\ref{fig:schematic}).

\emph{Context-copy redundancy} arises when the pipeline embeds the full
source narrative into the context field of every extracted item
(Figure~\ref{fig:schematic}). It dominates the two highest-volume
channels: the question-answering-with-reasoning (QAR) channel, where
81.0\% of 1.27 billion tokens are copied context, and the
relation-extraction (RE) channel, where 80.0\% of 732.6 million tokens
are copied context, as shown in Table~\ref{tab:per-task-redundancy}.
Because these channels emit many items per patient, each carrying its own
full copy of the narrative, a single narrative is reproduced on average
42.8 times within one patient's output, and up to 83 times for the most
complex presentations (Figure~\ref{fig:schematic}). The QAR channel alone
accounts for 1.03 billion copied-context tokens, more than the entire
272.6-million-token trainable-unique content of the corpus.

\emph{Generation-duplication redundancy} arises when the model produces
the same text repeatedly across patients, independent of any
context-copying. It dominates the risk-QA channel (51.3\% of tokens are
duplicated generated content), and is substantial in medication
extraction (25.9\%) and named-entity recognition (19.8\%). This redundancy reflects systematic regularities in instruction-tuned
model output. The model reuses a small set of phrasings, repeats similar
reasoning across records, and draws entity mentions from a bounded
vocabulary, so near-identical strings recur even when the underlying
patients differ. Unlike context-copy redundancy, it cannot be removed by
restructuring storage, because the repetition originates in the generation
process rather than in how records are assembled.

The distinction is consequential for mitigation. Context-copy redundancy
is eliminable by a storage change, referencing each narrative once per
patient rather than copying it per item, with no loss of information.
Generation-duplication redundancy can only be reduced by de-duplicating
the generated content itself, which discards genuine repeats.

\begin{table}[t]
\centering
\caption{\textbf{Per-task token composition and redundancy at full
167{,}034-patient scale.} For each of the ten text-bearing task channels, the table
reports total output tokens and their decomposition into copied context,
unique generated content, duplicated generated content, and scaffold
(identifiers/enumerated metadata). The two rightmost columns give the
percentage of each channel's tokens attributable to copied context
(context-copy redundancy) and to duplicated generation
(generation-duplication redundancy). The QAR and RE channels are
context-copy dominated; risk-QA, recommendations, risks, risk-states,
medication, and NER channels are generation-duplication dominated; the
summary channel is essentially clean. Token counts use the Llama-3.3-70B
tokenizer.}

\label{tab:per-task-redundancy}
\small
\setlength{\tabcolsep}{4pt}
\renewcommand{\arraystretch}{1.15}
\begin{tabular}{lrrrrrr}
\toprule
\textbf{Task} & \textbf{Total} & \textbf{Copied} & \textbf{Unique} &
\textbf{Dup.} & \textbf{\% ctx} & \textbf{\% dup} \\
 & \textbf{(M)} & \textbf{ctx (M)} & \textbf{gen (M)} &
\textbf{gen (M)} & \textbf{copied} & \textbf{gen} \\
\midrule
QAR              & 1267.9 & 1026.8 &  86.2 & 122.6 & 81.0 &  9.7 \\
RE               &  732.6 &  585.7 &   0.0 &  51.3 & 80.0 &  7.0 \\
Temporal events  &  156.5 &   73.8 &   0.0 &  32.8 & 47.1 & 20.9 \\
Risk-QA          &   56.8 &    0.0 &   2.0 &  29.1 &  0.0 & 51.3 \\
Summary          &   48.3 &    0.0 &  48.3 &   0.0 &  0.0 &  0.0 \\
NER              &   47.5 &    0.0 &  21.7 &   9.4 &  0.0 & 19.8 \\
Risk-states      &   44.4 &    8.9 &   7.6 &  20.4 & 20.1 & 46.0 \\
Recommendations  &   33.3 &    0.0 &   1.5 &  23.5 &  0.0 & 70.6 \\
Risks            &    9.9 &    0.0 &   0.0 &   6.8 &  0.0 & 68.1 \\
Medications      &    9.3 &    0.0 &   1.1 &   2.4 &  0.0 & 25.9 \\
\midrule
\textbf{All tasks} & \textbf{2510.7} & \textbf{1695.2} & \textbf{168.5} & \textbf{298.3} & \textbf{67.5} & \textbf{11.9} \\
\bottomrule
\end{tabular}
\end{table}

Channel-level mechanisms are heterogeneous: copied context dominates in
QAR and RE, whereas several downstream channels (e.g.\ recommendations
and risks) are driven primarily by duplicated generation. A visual
breakdown is provided in Supplementary Figure~\ref{fig:s1}; full values are
reported in Supplementary Table~\ref{tab:s2-mechanisms}.

\subsection{A reporting standard for synthetic-corpus information content}
The gap between apparent and actual content motivates a compact
reporting standard. Let $T_{\mathrm{total}}$ denote the total output
tokens of a corpus, partitioned by provenance into unique source text
$T^{\mathrm{u}}_{\mathrm{src}}$, unique generated content
$T^{\mathrm{u}}_{\mathrm{gen}}$, duplicated generated content
$T^{\mathrm{d}}_{\mathrm{gen}}$, copied context $T_{\mathrm{ctx}}$, and
scaffold $T_{\mathrm{scaf}}$, so that
$T_{\mathrm{total}} = T^{\mathrm{u}}_{\mathrm{src}} +
T^{\mathrm{u}}_{\mathrm{gen}} + T^{\mathrm{d}}_{\mathrm{gen}} +
T_{\mathrm{ctx}} + T_{\mathrm{scaf}}$. We define two ratios:
\begin{equation}
\mathrm{TUR} =
\frac{T^{\mathrm{u}}_{\mathrm{src}} + T^{\mathrm{u}}_{\mathrm{gen}}}
     {T_{\mathrm{total}}},
\qquad
\mathrm{CCR} = \frac{T_{\mathrm{ctx}}}{T_{\mathrm{total}}}.
\label{eq:metrics}
\end{equation}
The Trainable-Unique Ratio (TUR) is the fraction of output tokens
carrying information not already present elsewhere in the corpus; it
summarises how much of a corpus is informative. The Context-Copy Ratio
(CCR) is the fraction attributable to verbatim reproduction of source
context; it summarises how much is losslessly removable by a change in
serialisation. For the analysed corpus,
$\mathrm{TUR} = 0.109$ and $\mathrm{CCR} = 0.675$
(Equation~\ref{eq:metrics}, Table~\ref{tab:datacard}). Reported together with raw token count, they characterise a
synthetic redundancy profile of the corpus in a form that future releases
can adopt (Table~\ref{tab:datacard}).

Beyond informativeness, the redundancy carries direct storage, compute, and
downstream-processing costs that differ sharply between the two mechanisms;
we quantify these separately in Section~\ref{subsec:results-cost}.

\begin{table}[t]
\centering
\caption{Proposed reporting card for synthetic clinical corpora.
Values shown for the corpus analysed here. We recommend that
synthetic-corpus releases report these quantities alongside raw
token count, so that information content is not conflated with
volume.}
\label{tab:datacard}
\begin{tabular}{lr}
\toprule
\textbf{Reporting quantity} & \textbf{Value} \\
\midrule
Raw output tokens & 2.51\,B \\
Unique source tokens & 104.2\,M \\
Trainable-unique tokens & 272.6\,M \\
Redundant tokens & 1.99\,B \\
Scaffold tokens & 244.6\,M \\
\midrule
\textbf{Trainable-Unique Ratio (TUR)} & \textbf{0.109} \\
\textbf{Context-Copy Ratio (CCR)} & \textbf{0.675} \\
Redundancy ratio (vs.\ source) & 19.1$\times$ \\
\bottomrule
\end{tabular}
\end{table}

\subsection{The cost structure of redundancy}
\label{subsec:results-cost}
The two redundancy mechanisms impose costs of different kinds, and
separating them clarifies where each form of redundancy is costly.
We distinguish three costs, generation, storage, and downstream
processing, and show that copied-context and duplicated-generation
redundancy fall very differently across them
(Table~\ref{tab:cost}).

\paragraph{Generation compute is consumed only by duplicated generation}
Copied context is reproduced at serialisation time, not generated: the
model performs no additional inference to copy a source narrative into an
item's context field. Duplicated generation is different, each duplicate
string was produced by a forward pass through a 70-billion-parameter
model. Of the 466.8 million tokens of generated free text (unique plus
duplicated, excluding copied context and scaffold), 298.3 million, 64\% are
duplicates of strings the model had already produced. Nearly two-thirds of
the pipeline's free-text generation effort was therefore spent re-emitting
content it had generated before. This compute is unrecoverable: unlike
storage, it has already been spent, and it bears on the
approximately $1{,}200$ GPU-hours of generator inference the corpus required
(measured from the pipeline's job logs on NVIDIA H200 hardware), of which the
question-answering-with-reasoning and relation-extraction channels alone accounted for more
than half.

\paragraph{Storage is dominated by copied context.}
The text-bearing output occupies approximately 10\,GB on disk. Allocated
by token share, copied context accounts for roughly 7.5\,GB of this, with
duplicated generation contributing about 1.3\,GB and trainable-unique
content about 1.2\,GB. Three-quarters of the stored text is therefore
copied context that is losslessly removable by re-serialisation: storing
each source narrative once per patient and referencing it would reduce the
text footprint from $\approx$10\,GB to $\approx$2.5\,GB with no loss of
information.

\paragraph{Downstream processing scales with a 9.2$\times$ redundancy multiplier}
Every consumer that trains on, indexes, or re-processes the raw corpus
must read all 2.51 billion tokens to obtain the 272.6 million tokens of
trainable-unique content, a 9.2$\times$ processing multiplier on every
pass. This cost recurs each time the corpus is consumed, and it compounds
with corpus reuse: a corpus processed by many downstream studies pays the
multiplier once per study. Removing copied context alone lowers the
multiplier from 9.2$\times$ to 3.0$\times$; full de-duplication of
generated content lowers it further.

\begin{table}[t]
\centering
\caption{Cost structure of the two redundancy mechanisms. Copied-context
and duplicated-generation redundancy differ in which costs they impose:
copied context is cheap to generate but dominates storage and is losslessly
removable, whereas duplicated generation consumes generation compute that
cannot be recovered. All figures are derived from the measured per-category
token counts (Table~\ref{tab:per-task-redundancy}) and the corpus storage
and compute totals.}
\label{tab:cost}
\small
\renewcommand{\arraystretch}{1.25}
\begin{tabular}{lcc}
\toprule
 & \textbf{Copied context} & \textbf{Duplicated generation} \\
 & (1.70\,B tokens) & (298.3\,M tokens) \\
\midrule
Share of corpus            & 67.5\% & 11.9\% \\
Generation compute         & $\approx$\,none (copied) & real (64\% of free-text generation) \\
Storage footprint (text)   & $\approx$\,7.5\,GB & $\approx$\,1.3\,GB \\
Losslessly removable?      & Yes (re-serialise) & No (de-dup loses repeats) \\
\bottomrule
\end{tabular}
\end{table}

Taken together, the costs reinforce the two-mechanism distinction. Copied
context is a storage and processing cost that is fully recoverable;
duplicated generation is, in part, a compute cost that has already been
spent and cannot be undone, only avoided in future runs by less templated
generation.

\subsection{A clean channel shows redundancy depends on channel design}
The clinical-summary channel provides an internal control. Each patient
receives exactly one generated summary, and the summary does not embed
the source narrative. This channel contains 48.3 million tokens of which
0.0005\% are duplicated and 0\% are copied context: it is essentially
fully unique generated content. The per-channel composition in
Figure~\ref{fig:composition} makes this contrast immediate: the summary
channel is a single block of trainable-unique content, whereas QAR and RE
are dominated by copied context and the risk-derived channels by
duplicated generation. That this divergence occurs under an identical
pipeline, identical models, and identical corpus establishes that the
redundancy we measure is not an inevitable property of LLM-driven clinical
extraction. It depends on two structural properties of a channel: whether it emits many
context-bearing items per patient, and whether its outputs follow templated
schemas. A channel that emits one context-free generation per patient, as the
summary channel does, carries almost no redundancy; channels that ground many
items in source text carry more, by construction.

\subsection{Robustness to near-duplicate matching}
The unique-versus-duplicated distinction uses exact-match hashing, which
counts strings differing by a single token as distinct. To test whether
this materially undercounts redundancy, we applied MinHash near-duplicate
detection (Jaccard $>$ 0.85) to a 2\% sample of generated content
(76{,}400 strings). Exact matching marks 57.8\% of these strings as
unique; near-duplicate matching reduces this to 56.3\%, a difference of
only 1.5 percentage points (Supplementary Table~\ref{tab:neardup}). Restricting the relaxation to number-only
differences (e.g.\ two reasoning chains identical apart from a patient
age) accounts for just 0.2 points. The exact-match figures reported
throughout are therefore a conservative lower bound, and the
generated-content redundancy is predominantly verbatim repetition rather
than paraphrastic variation.

\subsection{Implications for clinical-data reproducibility and reuse}
The redundancy has three practical consequences for clinical
machine-learning practice.

First, corpus-size reporting is unreliable when stated in raw tokens. A
data statement describing this corpus as ``2.51 billion tokens''
overstates its information content ninefold relative to the 272.6 million
trainable-unique tokens. We recommend that synthetic clinical corpora
report unique-content size alongside raw size, analogous to de-duplication
reporting in web-scale pretraining corpora \citep{lee2022deduplicating}.

Second, training on the raw corpus risks distributional bias. A narrative
reproduced on average 42.8 times across the context fields of a patient's
items (Figure~\ref{fig:schematic}), which corresponds to a roughly sixteenfold
multiplication of source-\emph{token} volume because many copies are
sentence-level spans rather than full notes, contributes its content, its
conditions, demographics, and phrasings, that many times to any model trained
on the concatenated output. For clinical applications, where rare presentations are precisely
the cases of interest, silently up-weighting whichever narratives happen
to generate many items distorts the token-level training distribution of the
corpus, a distributional effect we measure directly, though the downstream
experiments below do not isolate a reliable rare-specific consequence of it.

Third, the redundancy is invisible at the point of use. Each output
record is individually well-formed; the duplication is only apparent in
aggregate. A consumer inspecting sample records would not detect that
nearly 80\% of the corpus is redundant. This motivates provenance-classification
of generated corpora as a routine quality-control step, for which we
release the classification tool used here.

\subsection{Compression independently confirms the redundancy}
\label{subsec:results-compression}
Lossless compression, applied with no knowledge of the provenance
categories, reproduces the PRD decomposition of
Section~\ref{subsec:results-mechanisms}. We compress four token streams, the
full corpus (\textsc{full}), its trainable-unique subset (\textsc{trainable}),
the copied-context stream (\textsc{copied-context}), and the duplicated-generation
stream (\textsc{dup-gen}), with Table~\ref{tab:compression} reporting their
compression ratios, computed on the full corpus; for the
dictionary and block-sorting compressors these agree with estimates from ten
10\% subsamples to within an absolute difference of $0.001$ in the ratio
(see below).

\begin{table}[t]
\centering
\caption{Compression ratio (compressed/raw; lower indicates more
redundancy) for the four token streams, \textsc{full} (entire corpus),
\textsc{trainable} (trainable-unique subset), \textsc{copied-context}, and \textsc{dup-gen} (duplicated generation), under four compressors, computed on the full corpus. The full corpus compresses
$2.7$--$4.7\times$ harder than the trainable-unique subset across all four compressor families.}
\label{tab:compression}
\small
\renewcommand{\arraystretch}{1.25}
\begin{tabular}{lcccc}
\hline
\textbf{Stream} & \textbf{gzip} & \textbf{bzip2} & \textbf{LZMA} & \textbf{PPMD} \\
\hline
\textsc{full}           & 0.059 & 0.053 & 0.031 & 0.043 \\
\textsc{trainable}      & 0.225 & 0.183 & 0.147 & 0.118 \\
\textsc{copied-context} & 0.035 & 0.042 & 0.021 & 0.037 \\
\textsc{dup-gen}        & 0.160 & 0.128 & 0.097 & 0.081 \\
\hline
\textbf{full vs.\ trainable} & 3.80 & 3.42 & 4.67 & 2.73 \\
\hline
\end{tabular}
\end{table}

\paragraph{The corpus compresses much better than its trainable core}
Across all four compressor families, the full corpus compresses
$2.7$--$4.7\times$ better than the trainable-unique subset
(Table~\ref{tab:compression}), and $3.4$--$4.7\times$ across the dictionary and block-sorting families. The trainable subset, with exact duplicates
and copied context removed, carries proportionally more information and
resists compression; the full corpus, dominated by repeated material, does
not. A compressor that knows nothing of the field taxonomy thus reaches the
same conclusion as PRD: most of the corpus is redundant.

\paragraph{The two mechanisms appear independently in compression}
The two losslessly-removable streams compress almost completely. The
copied-context stream reaches a compression ratio of $0.021$ under LZMA (a
$97.9\%$ reduction) and the duplicated-generation stream reaches $0.097$, both
far below the trainable-unique subset. A second, independent check compares how
much of the compressed corpus each mechanism occupies. Measured as a share of
the full corpus's compressed bytes, copied context accounts for $74.8\%$ and
duplicated generation for $12.9\%$; these byte shares track the token-level PRD
estimates ($67.5\%$ of tokens for copied context, $11.9\%$ for duplicated
generation). The byte and token fractions are not identical by construction,
because copied context consists of long contiguous spans that compress and pack
differently from short duplicated fragments, but the two independent methods
agree on both the overall magnitude of the redundancy and its split between the
two mechanisms.

\paragraph{The signal is stable under sampling and ordering}
The dictionary and block-sorting ratios are essentially insensitive to which
portion of the corpus is measured: across ten independent 10\% subsamples, every
such ratio has a standard deviation below $0.001$, and the full-corpus values
reported here differ from the subsampled means by at most $0.001$ (all on the
0--1 ratio scale, i.e.\ $0.1$ percentage points).

Shuffling record order before serialising changes the \textsc{full} ratio by at
most $0.0003$ across all four compressors, so the redundancy is a property of the
corpus contents rather than of their arrangement. The PPMD ratio is the one
exception to sampling-insensitivity: because its context model must observe both
members of a long-range duplicate pair to exploit them, it compresses
substantially better on the full corpus than on a 10\% subsample, which is why we
report full-corpus ratios throughout.

\paragraph{Context-modelling compression}
PPMD, the compressor whose context-modelling prediction is the closest classical
analogue to a language model's next-token prediction
\citep{teahan2018compression}, was run on the full corpus at model order 16. It compresses the full corpus to a ratio of $0.043$ (4.3\% of its original size), a full-versus-trainable gap of $2.73\times$, in line with the dictionary and block-sorting compressors. All four
families, spanning dictionary (gzip, LZMA), block-sorting (bzip2), and
context-modelling (PPMD) approaches, therefore agree on both the direction and
the approximate magnitude of the redundancy, which is what makes the
corroboration robust: a context model reaches the same conclusion as the
dictionary methods despite a wholly different mechanism. Compression here
validates the PRD decomposition; it is not itself the contribution.

\paragraph{Per-channel comparison: complementary, not identical}
Compression corroborates redundancy at the corpus level, but the two methods
measure related yet distinct quantities, which the per-channel comparison
makes explicit (Figure~\ref{fig:compression-perchannel}; full values in
Supplementary Table~\ref{tab:perchannel-supp}). PRD asks \emph{where each
token came from}, whether it was copied or duplicated, whereas compression
asks \emph{how predictable the text is}, regardless of origin. Channel-level
divergence between the two is therefore expected, not a discrepancy. The
methods agree most strongly on the channels richest in copied context:
question answering with reasoning (QAR) and relation extraction (RE), which PRD ranks as
most redundant ($90.7\%$ and $87.0\%$ respectively), are also the hardest to compress
($96.9\%$ and $96.3\%$ reduction). They diverge where text is stylistically regular
but not copied. The summary channel is the clearest case: PRD marks it as
carrying essentially no redundancy (it copies no source context and emits
one unique summary per patient), yet it still compresses by $81\%$, because
clinical summaries reuse similar phrasing and structure across patients even
when each is a distinct generation. The named-entity and medication channels
show the same pattern to a lesser degree. We therefore read the per-channel
result as complementarity rather than confirmation: PRD localises
\emph{where} redundancy originates, while compression additionally captures
\emph{stylistic predictability} that provenance, by design, does not mark.
The corpus-level agreement remains the relevant corroboration of the central
redundancy finding.

\subsection{De-duplication improves a downstream clinical encoder}
\label{subsec:results-downstream}
The redundancy analysis above is intrinsic: it measures the corpus without
reference to any model trained on it. To test whether the redundancy has a
\emph{downstream} consequence, we adapted a clinical encoder on the corpus and
evaluated it on two external, human-annotated benchmarks (NCBI-Disease and BC5CDR-Disease).

We continued masked-language-model pre-training of BioClinical
ModernBERT~\citep{sounack2025bioclinical} on three versions of the corpus,
each truncated to an identical token budget (174.3\,M whitespace tokens) so
that the conditions differ only in redundancy, not in training volume:
\textsc{raw} (the full corpus, all redundancy present), \textsc{dedup} (copied
context and duplicated generation removed, the trainable-unique subset), and
\textsc{ctx-removed} (copied context removed, generated duplicates retained;
an ablation isolating the copied-context mechanism). Each condition was adapted
under three random seeds, at a primary depth of 40{,}000 optimisation steps,
with 10{,}000- and 20{,}000-step adaptations run as a robustness ladder. We then froze each adapted encoder and trained a linear token-classification head on each
benchmark's training split, evaluating disease-NER entity F1 on its test set.
Both benchmarks are human-annotated and external to our pipeline, so they provide
an unbiased probe of the representations each condition learned. Test mentions
were stratified by their disease's frequency in our corpus into \textsc{rare},
\textsc{common}, and \textsc{unseen} (NCBI-Disease: $n=152$, $441$, $367$;
BC5CDR-Disease: $n=353$, $2{,}016$, $2{,}055$); the stratification was fixed
before training. Unless stated otherwise, the per-slice values quoted below are
for NCBI-Disease; the BC5CDR-Disease replication is reported at the end of this
section.

\begin{figure}[t]
\centering
\includegraphics[width=\linewidth]{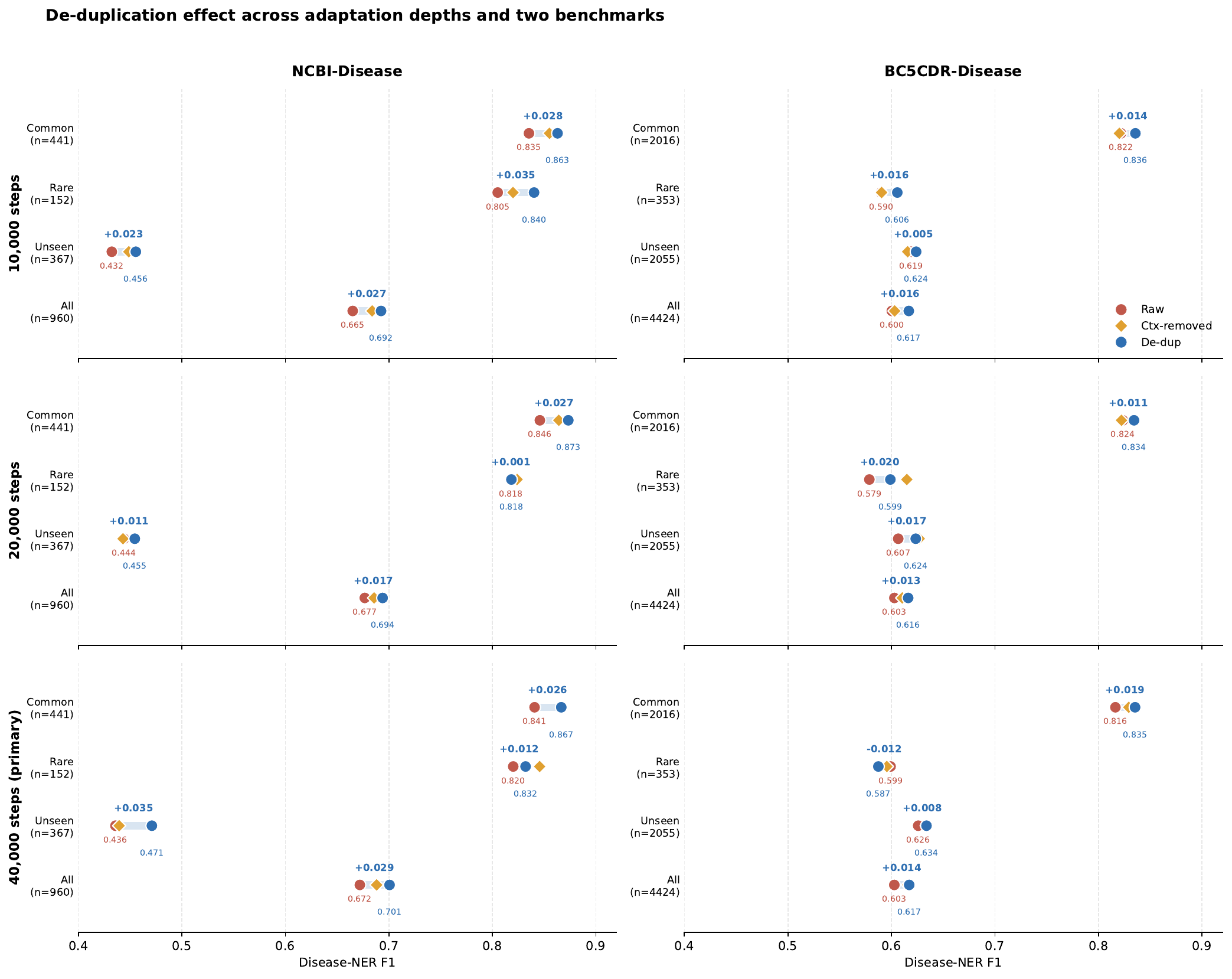}
\caption{\textbf{De-duplication improves a downstream clinical encoder across
adaptation depths and on two independent benchmarks.} Rows correspond to
adaptation depth (10{,}000, 20{,}000, and 40{,}000 primary steps, top to bottom);
columns to benchmark (NCBI-Disease \citep{dogan2014ncbi}, left; BC5CDR-Disease
\citep{li2016bc5cdr}, right). In each cell, for every frequency slice the band
spans the raw (red) to de-duplicated (blue) F1, with the context-removed ablation
(amber diamond) along it and the de-duplication gain annotated above; endpoint
values are shown below each band. Encoders were adapted by continued
masked-language pre-training of BioClinical ModernBERT
\citep{sounack2025bioclinical} on three versions of the corpus, each truncated to
an identical token budget. Reading down each column, the overall (All) and
common-disease gains are positive and stable across all three depths; reading
across each row, they reproduce on both benchmarks. The rare-slice effect is
small and does not reproduce: it varies in sign across depths (NCBI $+0.035$,
$+0.001$, $+0.012$; BC5CDR $+0.016$, $+0.020$, $-0.012$) and the
difference-in-differences between the rare and common slices is negative at the
deeper depths, consistent with the absence of a reliable rare-specific effect (on
NCBI-Disease the overall gain is significant by a mention-level bootstrap,
$p < 0.001$, whereas the rare-slice gain is not, $p = 0.67$).
Absolute F1 is not comparable across benchmarks, which cover different disease
distributions (corpus coverage $61.8\%$ NCBI, $36.1\%$ BC5CDR; the larger unseen
share on BC5CDR lowers its overall F1); the de-duplication \emph{gain} is the
comparable quantity. Bars are means over three seeds. The same adapted backbones
underlie both benchmarks; their adaptation loss is shown in Supplementary
Figure~\ref{fig:loss}, and full per-depth values are in
Table~\ref{tab:downstream} and Supplementary
Tables~\ref{tab:downstream-depths} and~\ref{tab:bc5cdr-depths}.}
\label{fig:downstream}
\end{figure}

\paragraph{De-duplication improves the encoder, robustly across adaptation depths}
At equal token budget, the encoder adapted on the de-duplicated corpus
outperformed the raw-corpus encoder overall and on every disease slice on
NCBI-Disease (Table~\ref{tab:downstream}, Figure~\ref{fig:downstream}): $+0.029$ F1 overall, $+0.026$ on common diseases, $+0.035$ on unseen, and $+0.012$ on rare diseases,
as means over three seeds at the primary adaptation depth (40{,}000 steps). The
overall and common-disease improvements were stable across adaptation depth: at
10{,}000 and 20{,}000 steps the overall gain was $+0.027$ and $+0.017$, and the
common-disease gain was $+0.028$ and $+0.028$, all in the same direction (Figure \ref{fig:downstream}; full per-depth values in Supplementary Tables \ref{tab:downstream-depths} and~\ref{tab:bc5cdr-depths}). Because the only difference
between conditions is whether the budget was spent on redundant or on unique
content, the result shows that redundant tokens carry less trainable signal than
unique ones: a corpus that is 79\% redundant trains a measurably weaker encoder
than its de-duplicated equivalent of the same size.

To confirm the overall effect is not an artefact of the three adaptation seeds,
we bootstrapped the de-duplication F1 gain over test mentions (2{,}000 resamples
of the 960 NCBI-Disease mentions, per-seed F1 averaged within each resample). The
overall gain is statistically significant (mean $+0.033$, $95\%$ CI
$[+0.018, +0.050]$, $p < 0.001$), as are the common- and unseen-disease gains
($p = 0.001$ and $p < 0.001$). The rare-slice gain is not significant
($+0.007$, $95\%$ CI $[-0.028, +0.041]$, $p = 0.67$), consistent with the absence
of a reliable rare-specific effect reported below.

\paragraph{Copied context accounts for most of the recoverable benefit}
The \textsc{ctx-removed} ablation, which removes only the copied-context
mechanism, recovered most of the improvement (overall F1 $0.688$, versus
$0.672$ raw and $0.701$ de-duplicated). The losslessly-removable copied-context
redundancy, the dominant component by volume
(Section~\ref{subsec:results-mechanisms}), thus also accounts for the larger
share of the recoverable downstream cost. This reinforces the practical
recommendation: eliminating copied context, a storage-level change that
discards no information, captures most of the available downstream benefit.

\paragraph{There is no reliable rare-specific benefit}
We had anticipated that redundancy might disproportionately harm rare
presentations, since the redundancy up-weights the item-rich common cases. The
data do not support this. The rare-slice gain was small and unstable across
adaptation depths ($+0.035$, $+0.001$, and $+0.012$ at 10{,}000, 20{,}000, and
40{,}000 steps), the rare-slice confidence intervals overlap between conditions
at every depth, and the difference-in-differences between the rare and common
slices, the rare-slice gain minus the common-slice gain, was negative at the two deeper depths ($-0.027$ and $-0.014$),
indicating that de-duplication benefits common diseases at least as much as
rare ones. Across three adaptation depths, we therefore find no reliable
rare-specific effect, and we report the downstream benefit of de-duplication as
a consistent overall improvement rather than a rare-specific one.

\paragraph{Single-source adaptation has a coverage cost}
On \textsc{unseen} diseases, those absent from our corpus, all three
conditions scored far lower than on seen diseases (on NCBI-Disease, F1
$0.44$--$0.47$ versus $0.82$--$0.87$). Adapting on a single-source corpus,
however large, does not confer recognition of diseases the corpus never
contained; the corpus covered $61.8\%$ of the distinct diseases in the
NCBI-Disease test set and $36.1\%$ in BC5CDR-Disease. This is a property of the
corpus's disease coverage, not of redundancy, and is only weakly affected by
de-duplication.

\paragraph{The corpus carries transferable clinical signal}
Independently of the redundancy comparison, the absolute performance is itself
informative: a frozen encoder adapted on the corpus and probed with only a
linear head reached $0.82$--$0.87$ F1 on the seen diseases of NCBI-Disease (and
$0.82$--$0.84$ F1 on common diseases in BC5CDR-Disease). The corpus, despite
being LLM-generated and heavily redundant, encodes clinical information that
transfers to human-annotated benchmarks, external evidence that its content is
clinically meaningful rather than an artefact of the generation pipeline.

\begin{table}[t]
\centering
\caption{Downstream disease-NER F1 (mean $\pm$ s.d.\ over three seeds) for
encoders adapted on the raw, de-duplicated, and context-removed corpus at equal
token budget (40{,}000-step primary adaptation), on two independent benchmarks.
On both benchmarks de-duplication improves the encoder overall and on common
diseases; the rare-slice effect is small and does not replicate (positive on
NCBI-Disease, negative on BC5CDR-Disease). The context-removed ablation recovers
most of the gain. Absolute F1 is not comparable across benchmarks (they cover
different disease distributions: corpus coverage $61.8\%$ NCBI, $36.1\%$ BC5CDR);
the de-duplication gain is the comparable quantity. The overall and
common-disease effects were stable across 10{,}000-, 20{,}000-, and
40{,}000-step adaptation (Figure~\ref{fig:downstream}; full per-depth values in
Supplementary Tables \ref{tab:downstream-depths} and~\ref{tab:bc5cdr-depths}). On NCBI-Disease, a mention-level bootstrap
($2{,}000$ resamples) confirms the overall gain is significant
($p < 0.001$); the rare-slice gain is not ($p = 0.67$).
}
\label{tab:downstream}
\footnotesize
\setlength{\tabcolsep}{4pt}
\renewcommand{\arraystretch}{1.15}
\begin{tabular}{llcccc}
\toprule
\textbf{Benchmark} & \textbf{Condition} & \textbf{Rare} & \textbf{Common} & \textbf{Unseen} & \textbf{All} \\
\midrule
\multirow{4}{*}{\shortstack[l]{NCBI-\\Disease}}
 & & ($n{=}152$) & ($n{=}441$) & ($n{=}367$) & ($n{=}960$) \\
 & Raw (redundant)        & $0.820 \pm 0.025$ & $0.841 \pm 0.010$ & $0.436 \pm 0.003$ & $0.672 \pm 0.003$ \\
 & De-duplicated          & $0.832 \pm 0.026$ & $0.867 \pm 0.009$ & $0.471 \pm 0.040$ & $\mathbf{0.701 \pm 0.019}$ \\
 & Context-removed (abl.) & $0.846 \pm 0.003$ & $0.867 \pm 0.018$ & $0.439 \pm 0.026$ & $0.688 \pm 0.019$ \\
\midrule
 & \textbf{\textit{De-dup.\ gain}} & $+0.012$ & $+0.026$ & $+0.035$ & $+0.029$ \\
\midrule
\multirow{4}{*}{\shortstack[l]{BC5CDR-\\Disease}}
 & & ($n{=}353$) & ($n{=}2016$) & ($n{=}2055$) & ($n{=}4424$) \\
 & Raw (redundant)        & $0.599 \pm 0.012$ & $0.816 \pm 0.013$ & $0.626 \pm 0.009$ & $0.603 \pm 0.004$ \\
 & De-duplicated          & $0.587 \pm 0.005$ & $0.835 \pm 0.007$ & $0.634 \pm 0.009$ & $\mathbf{0.617 \pm 0.002}$ \\
 & Context-removed (abl.) & $0.596 \pm 0.006$ & $0.829 \pm 0.009$ & $0.634 \pm 0.017$ & $0.616 \pm 0.001$ \\
\midrule
 & \textbf{\textit{De-dup.\ gain}} & $-0.012$ & $+0.019$ & $+0.008$ & $+0.014$ \\
\bottomrule
\end{tabular}
\end{table}

\paragraph{The effect replicates on a second benchmark}
To test whether the downstream effect generalises beyond a single benchmark, we
repeated the evaluation on BC5CDR-Disease~\citep{li2016bc5cdr}, an independent
disease-NER corpus, using the same frozen-backbone linear-probe protocol and the
same three adaptation depths (full results in Supplementary
Tables \ref{tab:downstream-depths} and~\ref{tab:bc5cdr-depths}). The pattern replicates: de-duplication improved
the encoder overall at every depth (gain $+0.016$, $+0.013$, $+0.014$ at
10{,}000, 20{,}000, and 40{,}000 steps) and on common diseases ($+0.014$,
$+0.011$, $+0.019$), in the same direction as on NCBI-Disease. The rare-slice
effect was again not reliable: the rare gain was small and changed sign across
depths ($+0.016$, $+0.020$, $-0.012$), and the difference-in-differences between
the rare and common slices was negative at the deepest depth ($-0.031$), despite
BC5CDR providing a larger rare slice ($n{=}353$) than NCBI-Disease ($n{=}152$).
The overall de-duplication gain is smaller on BC5CDR than on NCBI-Disease, and
the absolute F1 values differ because the two benchmarks cover different disease
distributions (the corpus covered $36.1\%$ of BC5CDR's distinct test diseases
versus $61.8\%$ of NCBI's); we therefore compare the de-duplication \emph{gain}
across benchmarks rather than absolute performance. Across two independent
benchmarks, the conclusion is consistent: de-duplication yields a measurable
overall improvement, concentrated in common diseases, with no reliable
rare-specific effect.

\section{Discussion}
A clinical machine-learning corpus is increasingly likely to be
synthetic: generated by a large language model rather than annotated by
clinicians. The volume such pipelines produce is routinely reported and
routinely impressive, billions of tokens, millions of extracted items.
Our central finding is that this volume is a poor proxy for information
content. In a production-scale clinical extraction pipeline, only 10.9\% of the
output tokens carried information not already present elsewhere in the
corpus, while 79.4\% were redundant and the remaining 9.7\% were
scaffold (identifiers and enumerated metadata). The corpus was, by raw
count, roughly an order of magnitude larger than it was by trainable-
unique content (a ninefold overstatement).

This gap matters because the raw count is what propagates into practice.
It is the number reported in data statements, used to estimate training
cost, and cited as evidence of scale. A clinical corpus described as
containing 2.51 billion tokens of multi-task extraction data conveys an
impression of richness that 272.6 million tokens of unique content does not
support. Prior work has shown that web-scale corpora contain enough
duplication, on the order of a few percent, to measurably harm the
models trained on them \citep{lee2022deduplicating}; the redundancy we
measure in synthetic clinical output is more than an order of magnitude
larger, and arises through a mechanism, per-item context copying, that
has no analogue in web-crawled data: web duplication reflects independent
re-publication of similar documents, whereas context-copy redundancy is
systematic intra-pipeline replication of the same source narrative across
an item set. Synthetic multi-task pipelines thus represent a more severe
redundancy regime than the web-scale case the field already takes
seriously.

\paragraph{Two mechanisms, two mitigations}
The redundancy we measured is not monolithic. Context-copy
redundancy, the verbatim reproduction of the source narrative inside
every extracted item, accounted for the bulk of it and is, in
principle, entirely eliminable: storing each narrative once per patient
and referencing it, rather than copying it per item, removes the
redundancy without discarding any information. Generation-duplication
redundancy, the model producing the same text across records, is
different in kind. It reflects the regularity of instruction-tuned
generation and can only be reduced by de-duplicating the generated content
itself, which removes genuine repeats. Distinguishing the two is
practically important: a practitioner who de-duplicates a corpus without
recognising that most of its redundancy is structural context-copying may
conclude that little can be saved, when in fact the largest component is
removable losslessly by a change to how the pipeline serialises its
output.

\paragraph{Origins of the two redundancy mechanisms}
The two redundancy mechanisms identified in section \ref{sec2} have distinct
origins, which determines how each should be interpreted and treated.
Context-copy redundancy is entirely deterministic and arises from the
pipeline's serialisation design: a verbatim copy of the source narrative
is attached to each extracted item so that both generator and verifier
operate with direct access to the underlying evidence. This mechanism is
architectural rather than emergent, and can be eliminated through
alternative representations (such as context pointers or shared
references) without altering the semantic content of the outputs.

Generation-duplication redundancy has a different character. The deployed
pipeline constrains generation to rigid output structures, specifically
the four-layer reasoning schema for the QAR channel and discrete schemas
for medication and temporal-event extraction, rather than permitting
free-form prose. Under such constraints, instruction-tuned LLMs exhibit a
well-documented regularity: a bounded output space encourages recurrent
phrasings, templated justifications, and a bounded entity vocabulary
across patient records. We do not isolate whether the observed regularity
is driven primarily by prompt design, schema rigidity, or intrinsic
decoding behaviour; distinguishing these would require controlled prompt
and model ablations, which we leave as future work. We therefore treat
generation-duplication as an observed property of the deployed system
rather than a fully attributed causal mechanism.

This regularity is not incidental but a designed trade-off. Constraining the
generator to schema-conformant, source-grounded outputs reduces the space in
which ungrounded content can be produced, supporting faithful, verifiable
extraction; the resulting redundancy is the cost of that grounding rather than
an inefficiency to be eliminated outright. Relaxing these constraints toward more variable
generation would likely decrease measured redundancy, but at the cost of
weakening the grounding the schemas enforce; the trade-off is thus
between token efficiency and output stability. Accordingly, the two
mechanisms call for different treatments: context-copy redundancy is
removable overhead and should be systematically eliminated, whereas
generation-duplication is a constraint-induced regularity that should be
reduced only where duplicate content can be removed without relaxing the
structural grounding that supports faithful extraction.

\paragraph{The cost of redundancy falls in two different places}
The two mechanisms are not only structurally but economically distinct, and
the distinction sharpens the case for treating them differently. Copied
context is cheap to produce, it is copied, not generated, but it dominates
storage (roughly three-quarters of the text-bearing corpus) and inflates every
downstream pass over the raw corpus by the 9.2$\times$ token multiplier; all of
this cost is recoverable by re-serialisation. Duplicated
generation is the opposite: it occupies far less storage, but it was
\emph{generated}, and 64\% of the pipeline's free-text generation effort
(298.3 of 466.8 million generated free-text tokens) was spent re-emitting
strings the model had already produced. That compute, unlike storage, is
already spent and cannot be recovered, only avoided in future runs by
relaxing the templated generation that produces it. The single most
actionable efficiency lever is therefore the cheaper one: eliminating copied
context costs nothing in information and removes the largest share of both
storage and downstream processing, whereas reducing generation-duplication
trades against the structural grounding the schemas provide.

\paragraph{The redundancy has a measurable downstream cost}
The redundancy is not only an accounting property of the corpus; it affects a
model trained on it. At equal token budget, an encoder adapted on the
de-duplicated corpus outperformed one adapted on the raw corpus on two
independent external disease-recognition benchmarks ($+0.029$ and $+0.014$ F1
overall on NCBI-Disease and BC5CDR-Disease at the primary adaptation depth,
stable in direction across three depths and replicated across both benchmarks;
Section~\ref{subsec:results-downstream}). Because the two conditions consumed
an identical number of training tokens, the difference is attributable to token
\emph{content}: redundant tokens carry less trainable signal than unique ones,
so a budget spent on a 79\%-redundant corpus trains a weaker model than the same
budget spent on its de-duplicated equivalent. The context-removed ablation
recovered most of the improvement, indicating that the losslessly-removable
copied-context mechanism, the larger component by volume, is also the larger
contributor to the downstream cost. This sharpens the practical message: the
single change that removes the most redundancy (eliminating copied context at
the serialisation layer) also captures most of the downstream benefit, at no
information loss. The benefit was a consistent overall improvement; we found no
reliable rare-specific effect, which did not replicate across the two benchmarks,
so the result converts the distributional concern into a measured overall
training cost rather than the rare-specific harm we had anticipated.

\begin{figure}[t]
\centering
\includegraphics[width=0.95\linewidth]{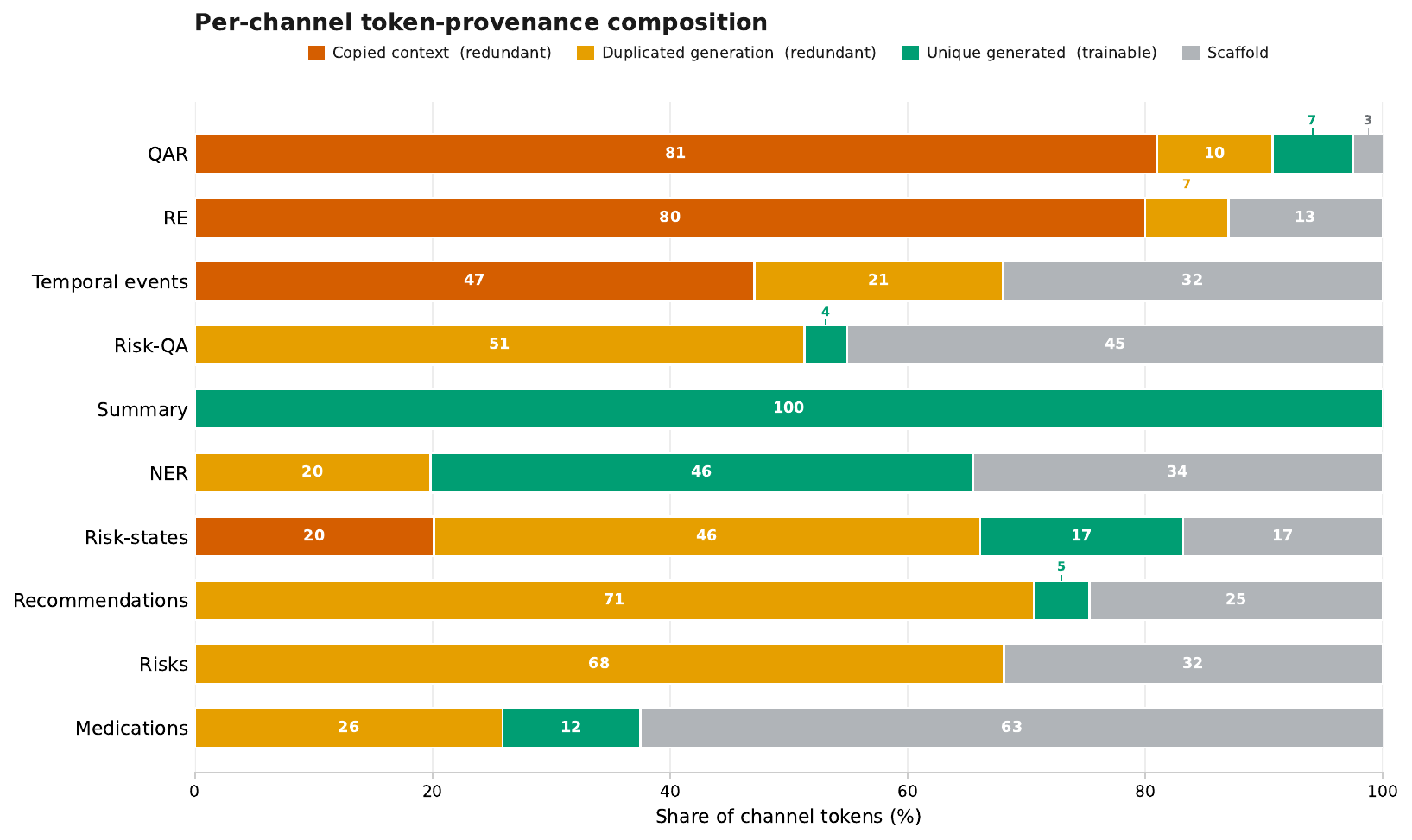}
\caption{\textbf{Per-channel token-provenance composition.} Each bar
decomposes one channel's output into copied context and duplicated
generation (both redundant), unique generated content (trainable), and
scaffold (identifiers and enumerated metadata); bars sum to 100\%. QAR and
RE are dominated by copied context; risk-QA, recommendations, risks, and
risk-states by duplicated generation. The summary channel is composed
entirely of unique generated content, demonstrating that redundancy
follows from channel design, emitting many context-bearing or templated
items per patient, rather than from LLM extraction itself.}
\label{fig:composition}
\end{figure}

\paragraph{The summary channel shows the redundancy is avoidable}
The summary channel is produced by the same models, on the same corpus,
through the same pipeline as every other channel, yet it contains almost
no redundancy (Figure~\ref{fig:composition}). The one thing that differs
is its design: the summary channel does not copy source context into its
output, and it emits one record per patient rather than many. Because the
redundant and non-redundant channels share everything except these design
choices, the redundancy in the other channels cannot be an inherent
property of LLM-driven extraction; it follows from how those channels are
structured.

\paragraph{Generality}
Our measurements come from one pipeline applied to one corpus, and the
specific proportions should not be assumed to transfer. The two
mechanisms, however, follow from design patterns common to LLM-driven
extraction pipelines rather than from anything specific to this corpus:
any pipeline that copies source context into per-item output will
accumulate context-copy redundancy, and any pipeline that emits free-text
generations across many items will accumulate generation-duplication
redundancy. We therefore expect the mechanisms to recur wherever these
patterns are present, while the exact proportions will depend on the
corpus, the task schema, and the degree of context copying. Measuring
these proportions across additional clinical corpora, including electronic
health record data with different documentation styles, is a natural next
study.

\paragraph{Why this matters for clinical models specifically}
The redundancy we measure is not a neutral accounting artifact; in a
clinical setting it bears on patient safety and equity. A synthetic corpus
whose described scale exceeds its information content invites overconfidence:
a model reported as trained on billions of clinical tokens may have seen an
order of magnitude less distinct clinical content, and downstream claims
about its competence inherit that overstatement. The consequence for a model
trained on the uncleaned corpus is twofold. First, duplication of the kind we
measure, both verbatim source copies and repeated generated strings, is
known to promote memorisation and degrade generalisation in language models
trained on it \citep{lee2022deduplicating}; an encoder trained on the raw
output expends capacity reproducing frequently repeated spans rather than
learning transferable clinical representations, and indeed we observe exactly
this, at equal token budget, de-duplication improves the adapted encoder on
two external benchmarks (Section~\ref{subsec:results-downstream}). Second, and
specific to the clinical setting, the redundancy is not uniform across
patients: because the records that spawn the most extracted items are the
longer, more complex presentations, uncorrected redundancy systematically
up-weights the patients whose narratives are richest, while the brief,
atypical presentation, the rare disease, the terse note, the
underdocumented patient, contributes proportionally less to the training
signal. A model trained without de-duplication would therefore be expected
to fit the common, heavily-replicated presentations more tightly while
under-representing the rare cases a clinical model most needs to recognise;
redundancy quietly tilts the learned distribution away from clinical
priority.

We measured the overall downstream consequence directly
(Section~\ref{subsec:results-downstream}): de-duplication improves the encoder
overall, robustly across adaptation depths. The rare-specific component of this
distributional concern, however, was not borne out: across three adaptation
depths and two independent benchmarks the rare-slice effect was small, unstable,
and statistically indistinguishable from the common-slice effect, so we report
the distributional skew as a measured property of the corpus whose specific
downstream harm to rare presentations remains, at the scale tested, a
well-motivated expectation rather than a demonstrated effect. Finally, because the duplication is
invisible in any single record and emerges only in aggregate, it escapes the
record-level review that clinical data governance typically relies on. These
are not reasons to abandon synthetic clinical corpora, which remain the only
practical route to data at this scale, but reasons to measure their
information content before trusting it, for which the provenance
classification we release provides a direct instrument.

\paragraph{Recommendations for practice}
Our findings translate into four concrete recommendations for the
production and release of synthetic clinical corpora.

First, report information content, not only volume. A corpus description
should state unique-content size alongside raw token count, and should
report the Trainable-Unique Ratio and Context-Copy Ratio
(Equation~\ref{eq:metrics}) so that consumers can distinguish the
informative content of the corpus from its apparent scale. The reporting card in
Table~\ref{tab:datacard} provides a template.

Second, separate the two redundancy mechanisms before acting. Because
context-copy redundancy is losslessly removable while
generation-duplication is not, a consumer should measure both
(for example with the provenance classifier we release) before deciding
how to de-duplicate. Treating all redundancy as a single quantity risks
either leaving the large removable component in place or discarding
genuine generated content.

Third, eliminate context-copy redundancy at the serialisation layer.
Storing each source narrative once per patient and referencing it,
rather than copying it into every extracted item, removes the dominant
redundancy component with no loss of information and no change to what
the pipeline generates. This is a change to corpus storage format, not
to the extraction model, and is therefore low-risk to adopt.

Fourth, de-duplicate generated content before training. The duplicated
generated component, though smaller, enters training data directly and
its removal follows established practice for reducing memorisation and
training-distribution bias in language models~\citep{lee2022deduplicating}.
In clinical applications this step also mitigates the over-representation
of common presentations relative to the rare cases of greatest clinical
interest.

Adopting provenance classification as a routine quality-control step
makes all four recommendations measurable rather than assumed, and
surfaces redundancy that is invisible to the record-level inspection
typically used to audit generated corpora.

\paragraph{Limitations}
Our analysis is conducted on a single pipeline applied to a single source
corpus, and the use of exact-match de-duplication (see Methods) renders the
reported redundancy estimates conservative. We assessed the downstream impact of
this redundancy directly (Section~\ref{subsec:results-downstream}): at equal
token budget, de-duplication improved an adapted clinical encoder overall and on
common diseases on two independent external benchmarks (NCBI-Disease and
BC5CDR-Disease), robustly across three adaptation depths.

Two factors bound the interpretation of this downstream result. First,
adaptation was performed using a single backbone (BioClinical ModernBERT) with a
linear probe; the magnitude of the effect may vary under full fine-tuning,
alternative encoder architectures, or longer continued pre-training. Second,
although we hypothesised that redundancy would disproportionately affect rare
presentations, the rare-slice effect was small and did not replicate across the
two benchmarks: it was not statistically significant at the primary depth
(bootstrap $p = 0.67$), varied in sign across adaptation depths, and the
difference-in-differences between the rare and common slices was negative at the
deeper depths. We therefore report a consistent overall improvement rather than a
rare-specific one, noting also that the rare/common stratification relied on
surface-string rather than ontology-level disease matching. Whether a
rare-specific effect emerges at larger scale or with semantic matching remains an
open question.

Finally, although the downstream effect now replicates across two independent
benchmarks, both are disease-entity recognition over PubMed abstracts, so they
share the academic-abstract genre and the disease entity type. Confirming that
the effect extends to the clinical-note genre that the pipeline ultimately
targets (for example the Problem/Treatment/Test concepts of the 2010 i2b2 task,
or disorder recognition on clinical notes as in ShARe/CLEF) and to broader entity
ontologies (for example MedMentions) is a natural direction for future work.

\paragraph{Conclusion}
The information content of a synthetic clinical corpus can be a small
fraction of its apparent size, and the shortfall is structured,
measurable, and, for its largest component, losslessly removable.
Reporting unique content, distinguishing the two redundancy mechanisms,
and de-duplicating before training or release are concrete, low-cost steps
that make synthetic clinical corpora more honestly described and more
fairly used.

\section{Methods}
\subsection*{Pipeline and corpus}
We analysed the complete output of a multi-agent clinical extraction
pipeline applied to the PMC-Patients corpus \citep{zhao2023large}, a
publicly released collection of 167{,}034 patient case-report narratives
drawn from PubMed Central full-text articles. The pipeline pairs a
generator model (Llama-3.3 70B) with a verifier model (MMed-Llama-3.1
70B) and emits structured outputs across eleven task channels, ten of which
emit free-text content and are the subject of this analysis: named-entity
recognition (NER), relation extraction (RE),
question-answering-with-reasoning (QAR), temporal-event extraction,
clinical summarisation, medication extraction, risk-grounded QA,
risk-based recommendations, risk-state derivation, and active-risk
identification. The remaining channel (visualisation-payload assembly)
emits only structured graph elements (nodes, edges, colours, levels) and
entity labels that are re-references of NER strings, with no newly
generated free text, and is excluded from the analysis. The three risk-derived channels
emit generated justification and reasoning text and, for risk-state
derivation, verbatim source-narrative slices in their evidence-anchor
fields; they are therefore within the scope of a token-provenance
analysis. Each patient record is stored as a single JSON object
containing the source narrative and the per-task outputs; the analysis
operates on the persisted output files exactly as written by the
pipeline, with no preprocessing.

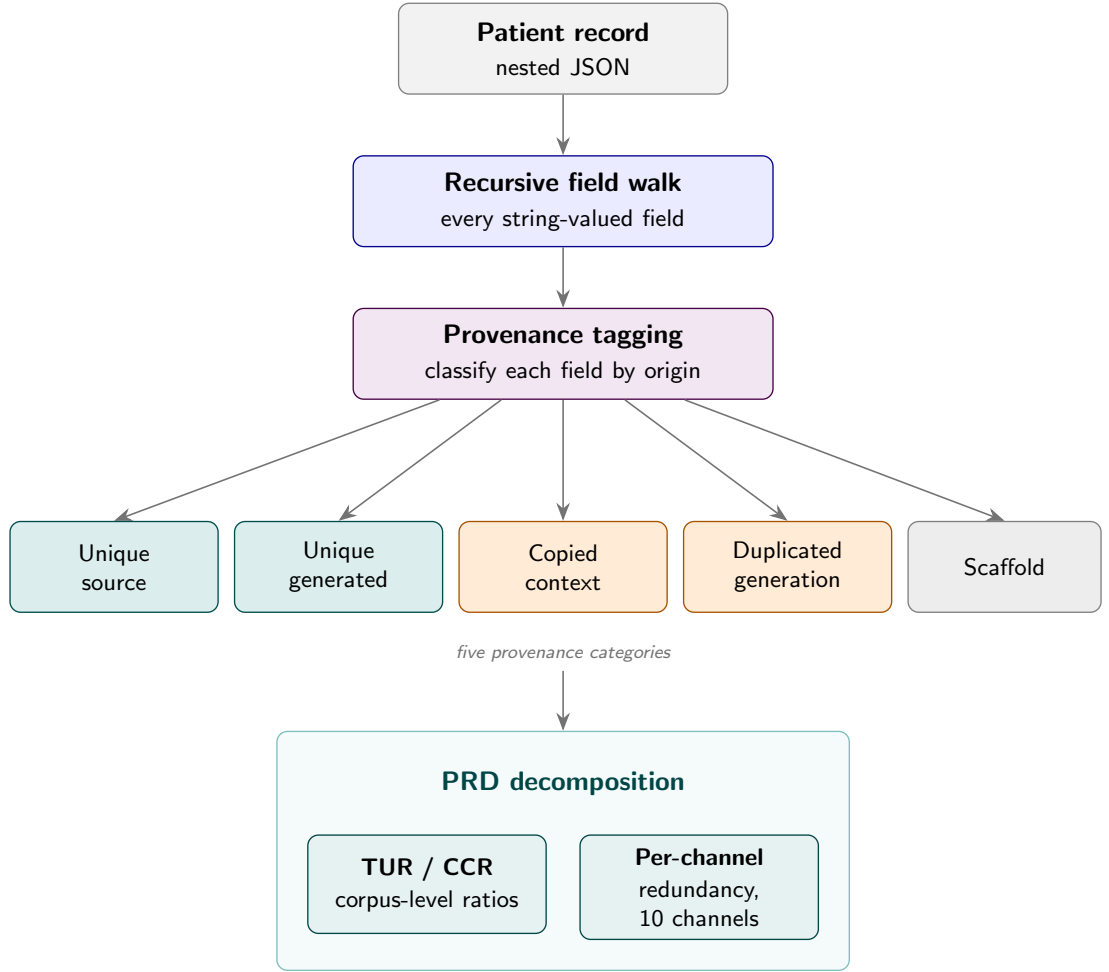
\begin{figure}[t]
\centering
\begin{tikzpicture}[
  font=\sffamily,
  node distance=7mm,
  >={Stealth[length=2.4mm]},
  box/.style={rounded corners=4pt, draw, line width=0.5pt,
              minimum height=12mm, inner sep=5pt, align=center},
  rec/.style={box,   fill=black!5,    draw=black!50,        text width=40mm},
  walk/.style={box,  fill=blue!8,     draw=blue!55!black,   text width=52mm},
  tag/.style={box,   fill=violet!10,  draw=violet!55!black, text width=52mm},
  uniq/.style={box,  fill=teal!14,    draw=teal!60!black,   text width=24mm,
               font=\sffamily\small},
  redbox/.style={box,fill=orange!16,  draw=orange!65!black, text width=24mm,
               font=\sffamily\small},
  scaf/.style={box,  fill=black!7,    draw=black!50,        text width=22mm,
               font=\sffamily\small},
  sub/.style={box,   fill=teal!10,    draw=teal!55!black,   text width=28mm,
               minimum height=13mm, font=\sffamily\small},
  arr/.style={->, draw=black!55, line width=0.6pt},
  txtlbl/.style={font=\sffamily\footnotesize\itshape, text=black!55},
]

\node[rec]  (rec)  {\textbf{Patient record}\\[1pt]\scriptsize nested JSON};
\node[walk] (walk) [below=8mm of rec]
  {\textbf{Recursive field walk}\\[1pt]\scriptsize every string-valued field};
\node[tag]  (tag)  [below=8mm of walk]
  {\textbf{Provenance tagging}\\[1pt]\scriptsize classify each field by origin};

\draw[arr] (rec)  -- (walk);
\draw[arr] (walk) -- (tag);

\node[redbox] (c3) [below=16mm of tag] {Copied\\context};
\node[uniq]   (c2) [left=2mm of c3]    {Unique\\generated};
\node[uniq]   (c1) [left=2mm of c2]    {Unique\\source};
\node[redbox] (c4) [right=2mm of c3]   {Duplicated\\generation};
\node[scaf]   (c5) [right=2mm of c4]   {Scaffold};

\foreach \c in {c1,c2,c3,c4,c5}{\draw[arr] (tag) -- (\c.north);}

\node[txtlbl] (catlab) [below=3mm of c3] {five provenance categories};


\node[font=\sffamily\bfseries, text=teal!55!black] (prd_title) [below=12mm of catlab.south] {PRD decomposition};

\node[sub] (tur) [below=4mm of prd_title, xshift=-18mm]
  {\textbf{TUR / CCR}\\[1pt]\scriptsize corpus-level ratios};
\node[sub] (perch) [below=4mm of prd_title, xshift=18mm]
  {\textbf{Per-channel}\\[1pt]\scriptsize redundancy, 10 channels};

\begin{scope}[on background layer]
  \node[box, fill=teal!4, draw=teal!50, fit=(prd_title)(tur)(perch), inner sep=4mm] (prd) {};
\end{scope}

\draw[arr] (catlab.south) -- (prd.north);

\end{tikzpicture}
\caption{\textbf{Overview of the provenance-based redundancy analysis.} Each
record's nested JSON object is recursively traversed and every string-valued field is
classified by origin into five provenance categories, trainable-unique
(unique source, unique generated), redundant (copied context, duplicated
generation), and structural scaffold. The provenance-based redundancy
decomposition aggregates the token counts into two outputs: corpus-level
ratios (the Trainable-Unique Ratio and Context-Copy Ratio) and a per-channel
redundancy breakdown across the ten text-bearing channels. An independent
compression analysis (Section~\ref{subsec:results-compression}) corroborates
the decomposition without using the provenance labels.}
\label{fig:method}
\end{figure}

\subsection*{Provenance-based Redundancy Decomposition (PRD)}
\label{subsec:methods-prd}
We quantified the information content of the generated corpus by a
procedure we term Provenance-based Redundancy Decomposition (PRD) (Figure~\ref{fig:method}):
classifying every output token into one of five provenance categories and
aggregating the redundant categories into corpus and channel-level
measures. Classification proceeds field by field over the nested JSON of
each record. For each string-valued field we assigned the category as
follows.

\textbf{Unique source text} The patient narrative (the top-level
\texttt{text} field) of each record. Source narratives are de-duplicated
across the corpus by exact-match hashing: each narrative is reduced to a
fixed-length MD5 fingerprint of its UTF-8 bytes, and narratives with
identical fingerprints are counted once, regardless of how many records
contain the narrative.

\textbf{Copied context} String fields whose function is to reproduce the
source narrative for the model's reference at extraction time, the
\texttt{context}, \texttt{verification\_anchor},
\texttt{verification\_ctx}, and \texttt{sentence} fields. These fields
embed the patient narrative verbatim (or a contiguous span of it) inside
individual extracted items and carry no information beyond the source
narrative already counted under \emph{unique source text}; their tokens
are therefore classified as redundant by construction. Where such a field
holds a placeholder rather than a narrative span (for example
``Context unavailable'' in a channel for which no context was attached),
it is classified as scaffold rather than copied context.

\textbf{Scaffold} Fields that carry structural metadata rather than
generated natural language: identifiers, character offsets, confidence
scores, and enumerated categorical values drawn from a closed vocabulary
(e.g.\ \texttt{verifier\_status}, \texttt{head\_type}, \texttt{tail\_type},
\texttt{label}, \texttt{category}, \texttt{event\_type}).

\textbf{Unique generated content} and \textbf{duplicated generated
content.} All remaining string fields, model-produced free text such as
answers, reasoning chains, summaries, and extracted entity strings. We
maintained a running set of MD5 hashes of previously observed generated
strings across the entire corpus; the first occurrence of a given string
is classified as unique generated content, and every subsequent identical
occurrence as duplicated generated content.

De-duplication is performed \emph{globally across the corpus}, not within
individual patients: a generated string is counted as unique on its first
occurrence anywhere in the 167{,}034-record corpus and as duplicated on every
subsequent identical occurrence, regardless of which patient produced it.
Source narratives are likewise de-duplicated against a corpus-wide set.
Copied-context tokens are not de-duplicated; every per-item copy of a source
span is counted as redundant by construction, since each copy is the
mechanism of context-copy redundancy.

The field-to-category assignment was fixed in advance from the pipeline's
output schema and applied uniformly across all records; The complete field-to-category mapping is given in Supplementary Table~\ref{tab:fieldmap} and released with the analysis tool, so the classification is fully specified and reproducible.

\subsection*{Compression as a model-free redundancy check}
\label{subsec:methods-compression}
Provenance-based Redundancy Decomposition
(Section~\ref{subsec:methods-prd}) quantifies redundancy by attributing
every token to a source. To corroborate that measurement with an
independent signal, we additionally estimate redundancy through classical
lossless compression. Compression ratio is a long-established proxy for
statistical redundancy: a stream that compresses to a small fraction of its
size is, by construction, highly predictable from its own contents
\citep{cleary1984ppm}. Crucially, compression uses none of the PRD
provenance categories and no knowledge of the corpus schema, so agreement
between the two methods is not circular. Compression captures statistical
predictability regardless of origin, whereas provenance explicitly tracks
source attribution; the two are therefore expected to align only where
redundancy arises from repeated content, and to diverge where text is
predictable for other reasons.

We serialise the corpus into four byte streams using the \emph{same}
recursive field walk and the \emph{same} provenance taxonomy as PRD, so the
two analyses cover identical text, including text nested inside structured
fields: (i) \textsc{full}, every text-bearing field in corpus order;
(ii) \textsc{trainable}, the deduplicated unique source narratives plus the
first occurrence of each generated string (the trainable-unique subset);
(iii) \textsc{copied-context}, the copied-context fields alone; and
(iv) \textsc{dup-gen}, the second-and-later occurrences of generated
strings. Streams (iii) and (iv) isolate the two redundancy mechanisms of
Section~\ref{subsec:results-mechanisms}. Each stream is compressed with four
compressors spanning three algorithmic families: gzip (LZ77 with Huffman
coding) \citep{ziv1977}, bzip2 (Burrows--Wheeler transform)
\citep{burrows1994}, LZMA, and PPMD (prediction by partial matching)
\citep{cleary1984ppm, teahan2018compression}. PPM is the family whose
context-modelling prediction is the closest classical analogue to a
language model \citep{teahan2018compression}, making its compression ratio
a particularly apt redundancy measure for model-generated text.

We compress each stream on the full corpus. The full text-bearing corpus
(approximately 10\,GB) is processed in a two-stage memory-bounded procedure: the
four provenance streams are first serialised to disk, then each is compressed
sequentially, so that LZMA and PPMD run at full scale within memory. PPMD is run
at model order 16 (the context length over which it models token statistics) with
2048\,MB of working memory. As a robustness check, we also estimated each ratio by
Monte-Carlo subsampling, drawing ten independent 10\% subsamples
($\approx 16{,}700$ records each): for the dictionary and block-sorting
compressors, the subsampled means matched the full-corpus ratios to within
$0.001$, confirming the estimate is stable; PPMD, whose context model must observe
both members of a long-range duplicate pair to exploit it, compresses
substantially better on the full corpus than on a subsample, which is why we
report full-corpus ratios throughout. We additionally compress a record-shuffled
\textsc{full} stream to confirm the result does not depend on record ordering. Compression serves only as an independent validation
of the provenance decomposition; it does not define redundancy in our framework, which is established by token provenance (Section~\ref{subsec:methods-prd}).

\subsection*{Downstream evaluation: does redundancy affect a trained encoder?}
\label{subsec:methods-downstream}
To test whether the measured redundancy has a downstream consequence, we
adapted a clinical encoder on the corpus and evaluated it on two external,
human-annotated benchmarks. The design was fixed before any model was trained.

\paragraph{Conditions and equal-budget control}
We constructed three training corpora from the pipeline output using the same
provenance taxonomy as PRD (Section~\ref{subsec:methods-prd}):
\textsc{raw} (every text-bearing field, all redundancy present),
\textsc{dedup} (de-duplicated unique source narratives plus the first
occurrence of each generated string, the trainable-unique subset), and
\textsc{ctx-removed} (copied context removed, generated duplicates retained;
an ablation isolating the copied-context mechanism). Because the raw corpus
contains roughly $8.5\times$ more tokens than the de-duplicated corpus when both
are counted in whitespace tokens on the constructed training corpora (the
$9.2\times$ provenance multiplier reported above is computed on Llama-tokenizer
tokens over the full corpus; the two differ because of tokenizer and
field-selection differences, not measurement error), each
corpus was truncated to an identical budget of $174.3$\,M whitespace tokens, so
that the conditions differ only in the \emph{kind} of tokens (redundant versus
unique), not their number. This removes training volume as a confound.

\paragraph{Continued pre-training}
Each corpus was used to continue masked-language-model pre-training of
BioClinical ModernBERT-base \citep{sounack2025bioclinical}, a $150$\,M-parameter
encoder pre-trained on biomedical and clinical text. We selected this backbone
because it is a current, clinically-specialised encoder built on the ModernBERT
architecture \citep{warner2025smarter}, which delivers a state-of-the-art
performance-efficiency trade-off among encoder-only models while remaining a
masked-language-model encoder suitable for continued pre-training and
linear-probe evaluation. A clinically pre-trained backbone is the appropriate
substrate for this experiment, and because the comparison is
\emph{within-backbone}, the same encoder adapted on each corpus condition at
equal budget, the choice of backbone sets the starting point but does not affect
the de-duplication contrast we measure, which depends only on the difference
between conditions. Adaptation used a masking
probability of $0.15$, batch size $32$, maximum sequence length $256$, learning
rate $5\times10^{-5}$ with $200$ warmup steps, and $40{,}000$ optimisation steps
as the primary adaptation depth, under mixed-precision training. We additionally ran $10{,}000$- and $20{,}000$-step adaptations as a robustness
ladder; the stability of the downstream effect across all three depths is shown
in Figure~\ref{fig:downstream}, and the corresponding training-loss trajectories
in Supplementary Figure~\ref{fig:loss}. Each condition was adapted under three
random seeds, yielding nine condition-specific backbones per depth, on which both downstream benchmarks were probed.

\paragraph{Probing protocol}
To measure the representations each condition learned rather than the capacity
of a fine-tuned model, we used a linear probe: each adapted backbone was frozen
and a single token-classification head was trained on the benchmark's training
split, with disease-NER entity F1 evaluated on its test split. We applied this
identical protocol to two independent, human-annotated benchmarks external to
our generation pipeline: NCBI-Disease \citep{dogan2014ncbi} ($941$ test
sentences, $960$ disease mentions) and BC5CDR-Disease \citep{li2016bc5cdr}
($5{,}865$ test sentences, $4{,}424$ disease mentions; the chemical annotations
were collapsed to the outside class so the task matches NCBI's disease-only
scheme). Because both benchmarks are human-annotated and external, they provide
an unbiased probe and avoid circularity. The probe was trained for five epochs
(learning rate $10^{-3}$, maximum sequence length $256$); the procedure was
identical across both benchmarks, all conditions, and all seeds. Because the
benchmark probe is independent of corpus adaptation, both benchmarks were probed
on the \emph{same} adapted backbones, so the two evaluations share the
adaptation but differ only in the downstream benchmark.

\paragraph{Frequency stratification}
To test whether any effect concentrated on rare diseases, we stratified each
benchmark's test mentions by the frequency of their disease in our training
corpus. Each distinct test disease was matched to the corpus by normalised
surface string (lower-cased, whitespace-collapsed); this matching is
surface-level, not ontology-normalised, because the corpus's extracted diagnoses
are not mapped to a controlled vocabulary. A mention was labelled \textsc{rare}
if its disease fell in the bottom quartile of the seen-disease frequency
distribution, \textsc{common} if above it, and \textsc{unseen} if its disease
never appeared in the corpus. The same bottom-quartile rule was applied to each
benchmark, yielding a dataset-relative threshold: corpus frequency $\leq 3$ for
NCBI-Disease and $\leq 30$ for BC5CDR-Disease. For NCBI-Disease the corpus
contained $194$ of the $403$ distinct test diseases ($61.8\%$ coverage), giving
$441$ common, $152$ rare, and $367$ unseen mentions; for BC5CDR-Disease it
contained $483$ of $1{,}337$ distinct test diseases ($36.1\%$ coverage), giving
$2{,}016$ common, $353$ rare, and $2{,}055$ unseen mentions. The stratification
rule and the quartile threshold were fixed from coverage statistics before any
model was trained.

\paragraph{Statistics}
We report mean and standard deviation of entity F1 across the three seeds per
condition and slice, with $95\%$ confidence intervals from $1{,}000$ bootstrap
resamples of the test mentions (per-slice CIs use $1{,}000$ resamples; the
overall-gain significance test below uses $2{,}000$, a deliberate difference,
not a typo). The headline comparison is the per-slice
difference between the de-duplicated and raw conditions, and the
difference-in-differences between the rare and common slices. Both quantities
are reported for each benchmark. To test whether the overall de-duplication gain
is robust beyond the three adaptation seeds, we additionally bootstrap the gain
over test mentions ($2{,}000$ resamples, per-seed F1 averaged within each
resample) and report a two-sided $p$-value. Mention-level bootstrap significance
is reported for NCBI-Disease, for which per-mention predictions were retained;
for BC5CDR-Disease we report the de-duplication gain and its direction across
depths.

\subsection*{Validation of the field-to-category mapping}
To confirm that the provenance categories reflect the actual content of
each field rather than only its schema label, we validated the mapping on
a random sample of 200 records. For every field classified as copied
context (the \texttt{context} and \texttt{sentence} fields), we tested
whether its content was a verbatim substring of the record's source
narrative: all 9{,}031 such fields (100.0\%) were exact source
substrings, confirming the copied-context assignment. For fields
classified as generated content (\texttt{answer}, \texttt{reason}, and
\texttt{summary}), we tested whether they were absent from the source;
2{,}196 of 2{,}289 (95.9\%) were novel text not present in the narrative.
The 4.1\% that did appear in the source were exclusively short extractive
\texttt{answer} spans, in which the model's answer quotes a phrase
verbatim from the narrative (for example, ``discharged on oral phenytoin'' or ``died due to respiratory failure''). These are correct
extractions rather than misclassifications, and because they are counted
as generated content they make the trainable-unique total marginally
larger and the redundancy estimate correspondingly conservative. The
validation script is released with the analysis tool.

\subsection*{Derived quantities}
From the per-category token counts, we computed: \emph{trainable-unique
content} (unique source text $+$ unique generated content);
\emph{redundant content} (copied context $+$ duplicated generated
content); and the \emph{redundancy ratio} (redundant content divided by
unique source text). Per task, we report the total token count and its
decomposition into copied context, unique generated, duplicated
generated, and scaffold, together with the proportion of each task's
tokens attributable to copied context, distinguishing the two redundancy
mechanisms.

\subsection*{Tokenisation}
All token counts were computed with the tokenizer of the pipeline's
generator model (Llama-3.3 70B), so that the reported figures correspond
to the token budget the corpus would consume in a model of that family.
Token counts are tokenizer-dependent; counts under a different tokenizer
would differ in absolute magnitude but not in relative composition, since
all categories are tokenised identically.

\subsection*{Implementation and reproducibility}
The analysis was implemented as a single-pass streaming classifier over
the pipeline's output files and run over the complete 167{,}034-record
corpus. The classifier, the field-to-category mapping, and the script
that produced all reported figures are released openly to permit exact
replication on this corpus and application to other multi-task extraction
pipelines (as shown in Supplementary Table~\ref{tab:config}).

\section*{Data and code availability}
The token-provenance classifier, its validation and near-duplicate
robustness checks, the figure-generation scripts, and the aggregate
per-channel token counts that underlie every headline figure in this paper are
openly available on GitHub at
\url{https://github.com/Ali-Lazem/clinical-corpus-redundancy} and permanently
archived on Zenodo at \url{https://doi.org/10.5281/zenodo.20848582}. Releasing
the aggregate per-channel counts allows the redundancy decomposition to be
verified independently of the raw corpus. The extraction pipeline that produced
the corpus, and the derived corpus itself, are not included; the corpus can be
regenerated from the publicly available PMC-Patients narratives
\citep{zhao2023large} using the released analysis code.

\begin{figure}[t]
\centering
\includegraphics[width=\linewidth]{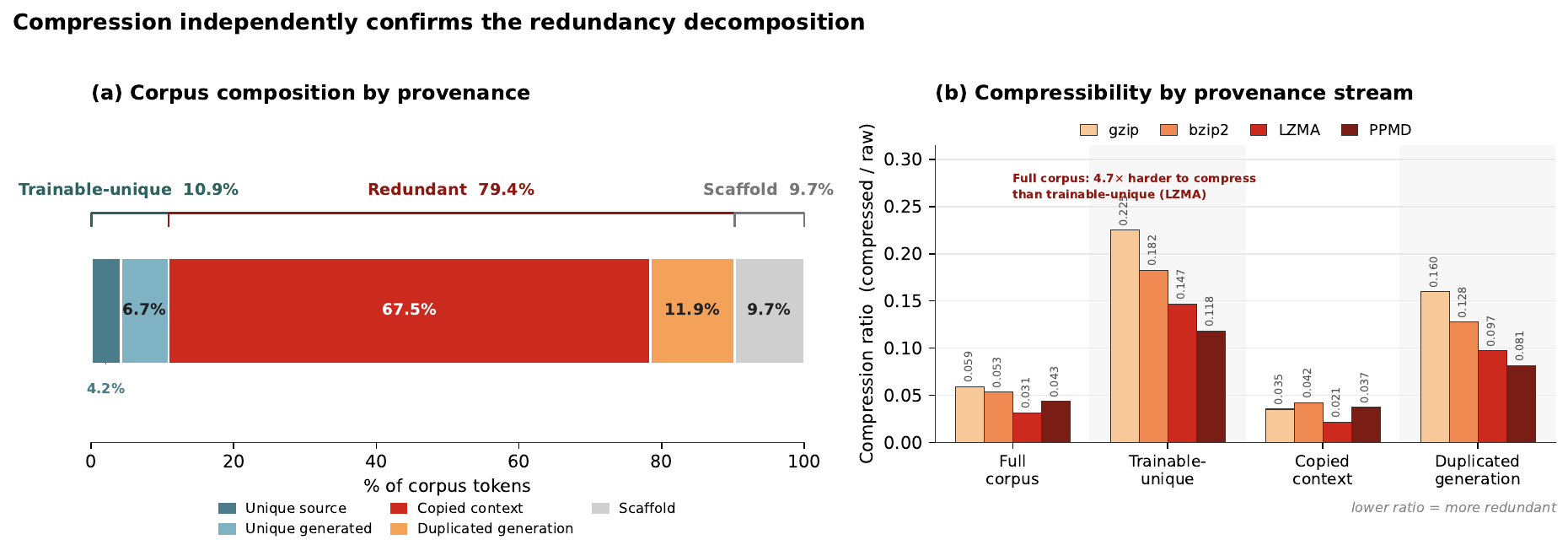}
\caption{\textbf{Compression independently confirms the redundancy
decomposition.} \textbf{(a)} Corpus composition by token provenance: the
2.51-billion-token output partitioned into the five provenance categories.
Unique source (4.2\%) and unique generated (6.7\%) tokens together form the
trainable-unique content (10.9\%); copied context (67.5\%) and duplicated
generation (11.9\%) form the redundant content (79.4\%); the remainder is
structural scaffold (9.7\%). \textbf{(b)} Compressibility of each provenance
stream under four lossless compressors spanning three algorithmic families,
computed on the full corpus (compression ratio = compressed/raw; lower indicates
more redundancy). The trainable-unique subset is the only stream that resists
compression; the full corpus, copied context, and duplicated generation all
compress to a small fraction of their size. The full corpus compresses
$2.7$--$4.7\times$ more than the trainable-unique subset across all four
compressor families (equivalently, the trainable-unique subset is
$2.7$--$4.7\times$ harder to compress). Compression uses no knowledge of the provenance
categories, so its agreement with panel~(a) is an independent confirmation of the
decomposition.}
\label{fig:compression}
\end{figure}

\begin{figure}[t]
\centering
\includegraphics[width=\linewidth]{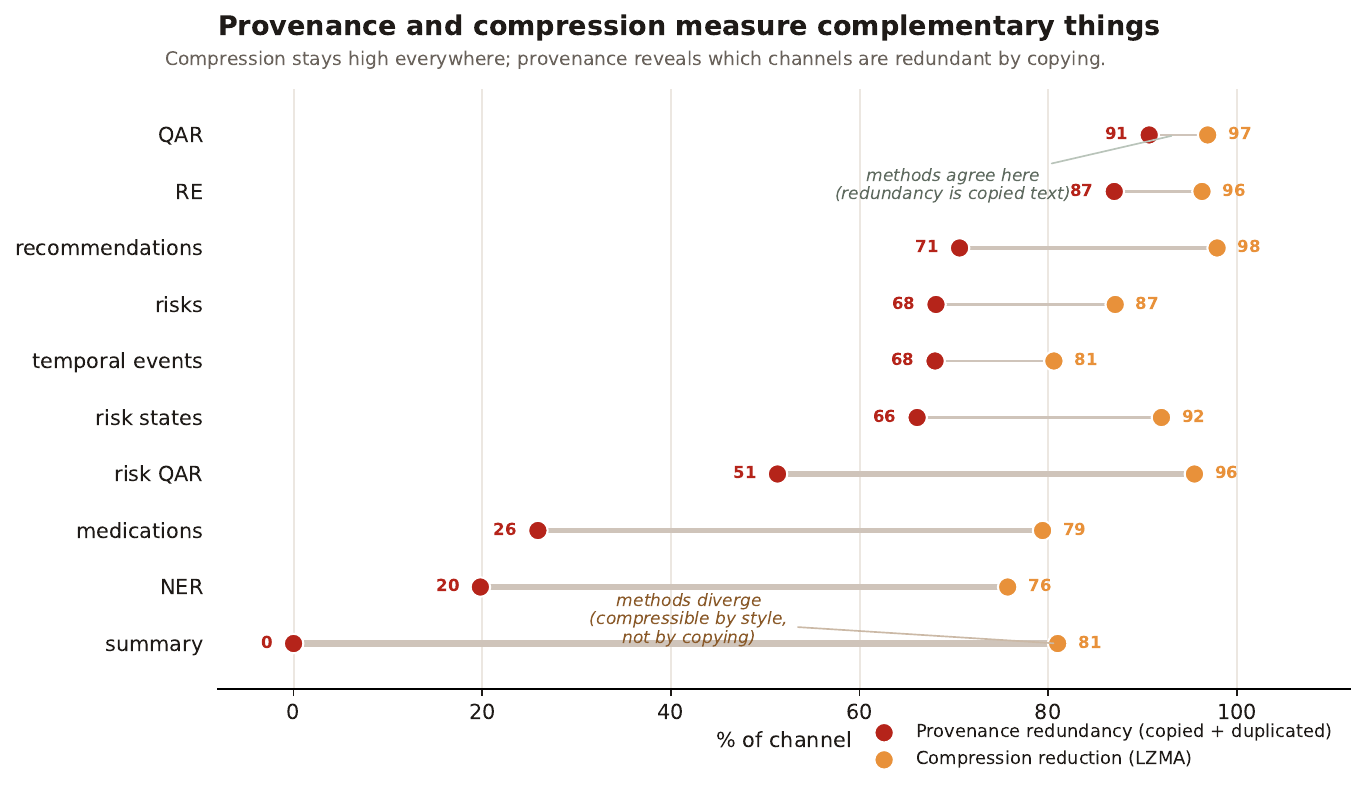}
\caption{\textbf{Provenance and compression measure complementary properties of
each channel.} For every channel, the deep-red marker is the provenance
redundancy (copied-context plus duplicated-generation, as a percentage of the
channel's tokens) and the amber marker is the compression reduction (LZMA); the
connecting bar is the gap between them. The two methods agree on the channels
richest in copied context (QAR, RE), which rank highest on both. They diverge
where text is stylistically regular but not copied: the summary channel, which
provenance marks as carrying no copied redundancy, still compresses by $81\%$
because clinical summaries reuse similar phrasing across patients. Provenance
localises \emph{where} redundancy originates; compression additionally registers
stylistic predictability. Channels ordered by provenance redundancy; full values
in Supplementary Table~\ref{tab:perchannel-supp}.}
\label{fig:compression-perchannel}
\end{figure}

\section*{Author contributions}
A.H.L. conceived the study, performed the analysis, and wrote the
manuscript. W.T. supervised the work and revised the manuscript.

\section*{Acknowledgements}
We acknowledge the services of Supercomputing Wales and the Bangor
eResearch Team, including the support provided by Dr. Ade Fewings.

\section*{Funding}
This work was supported by the Ministry of Higher Education and Scientific
Research in Iraq.

\section*{Competing interests}
The authors declare no competing interests.

\bibliography{sn-bibliography}
\clearpage 

\newcolumntype{R}{>{\raggedleft\arraybackslash}X}
\newcolumntype{C}{>{\centering\arraybackslash}X}

\begin{appendices}
\setcounter{table}{0}
\setcounter{figure}{0}
\renewcommand{\thetable}{S\arabic{table}}
\renewcommand{\thefigure}{S\arabic{figure}}
\renewcommand{\theHtable}{S\arabic{table}}
\renewcommand{\theHfigure}{S\arabic{figure}}

\section{Supplementary Information}\label{secSI}

\subsection{Overview}
\begin{figure}[htbp]
\centering
\includegraphics[width=\textwidth]{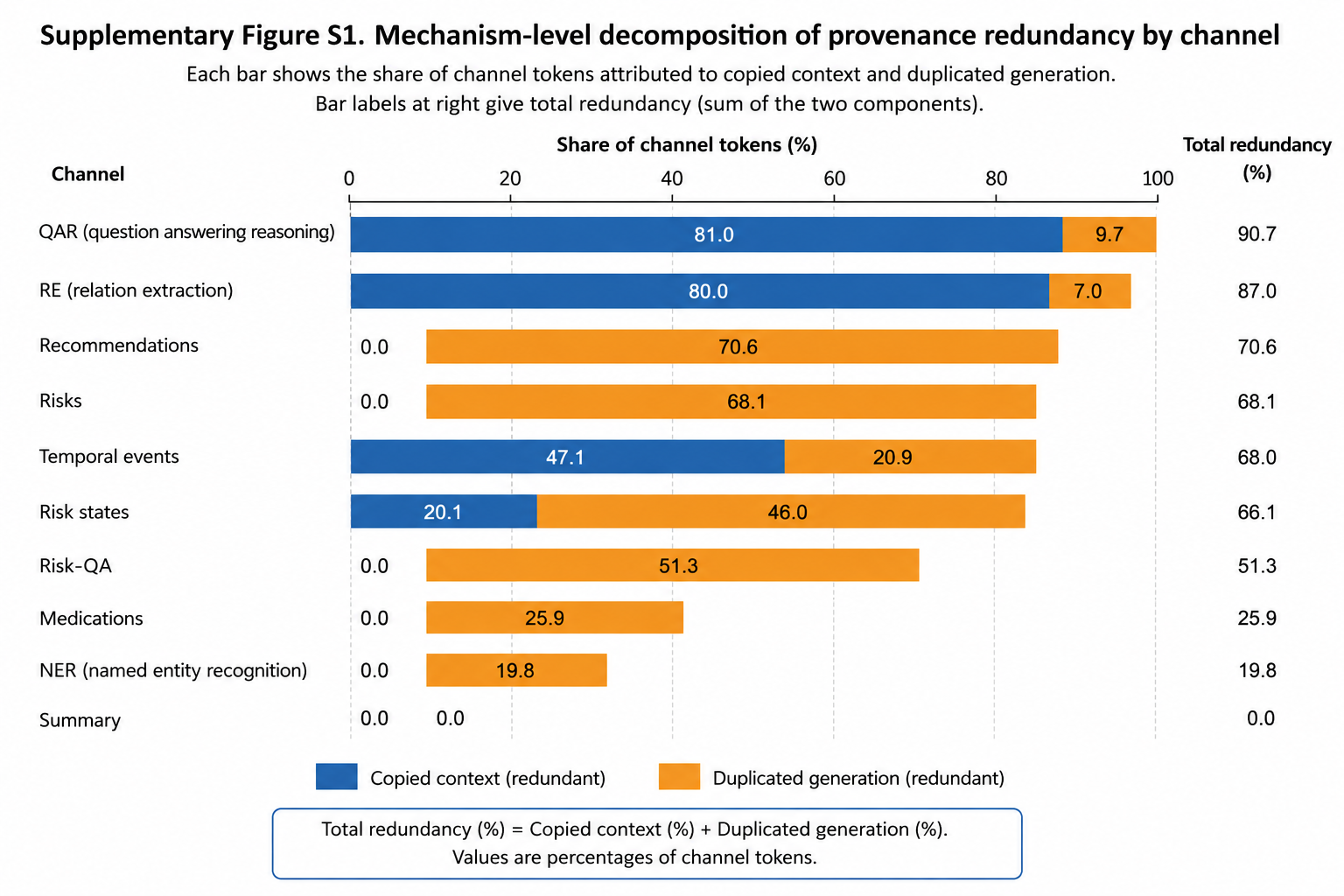}
\caption{Mechanism-level decomposition of provenance redundancy by channel.
Each bar is split into copied context (blue) and duplicated generation
(orange); the total redundancy is their sum, shown at right. Channels
dominated by copied context (QAR, RE) exhibit near-verbatim reuse of source
material, whereas the risk-derived channels (recommendations, risks, risk-QA)
are dominated by duplicated generation across records. The summary channel
contains neither copied nor duplicated content, yet remains compressible
(Supplementary Table~\ref{tab:perchannel-supp}) owing to shared linguistic
structure across patients. This decomposition shows that redundancy is not a
single phenomenon but arises through two mechanisms that dominate different
channels, which is why corpus token counts overstate trainable content.}
\label{fig:s1}
\end{figure}

\begin{figure}[htbp]
\centering
\includegraphics[width=\linewidth]{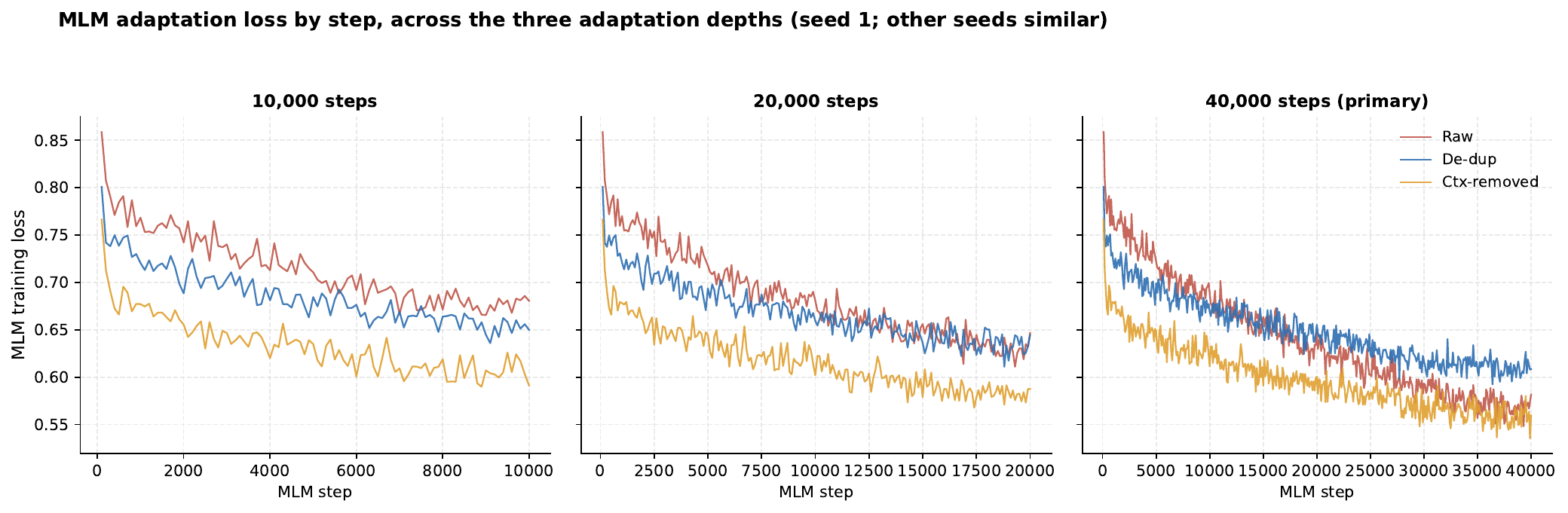}
\caption{\textbf{MLM adaptation loss across the three adaptation depths}
(seed~1; other seeds are quantitatively similar). Each panel shows the training
loss for the raw, de-duplicated, and context-removed conditions at the
corresponding adaptation depth. The loss falls steeply over the first few
thousand steps and progressively flattens; by the 40{,}000-step primary depth
the raw and de-duplicated conditions that carry the principal comparison have
substantially plateaued (last-1000-step change $<0.006$), while the
context-removed ablation continues a slow descent. The downstream effect is
stable in direction across all three depths
(Figure~\ref{fig:downstream}), indicating the comparison is not an artefact of a particular adaptation depth.}
\label{fig:loss}
\end{figure}

\begin{figure}[htbp]
\centering
\includegraphics[width=0.95\linewidth]{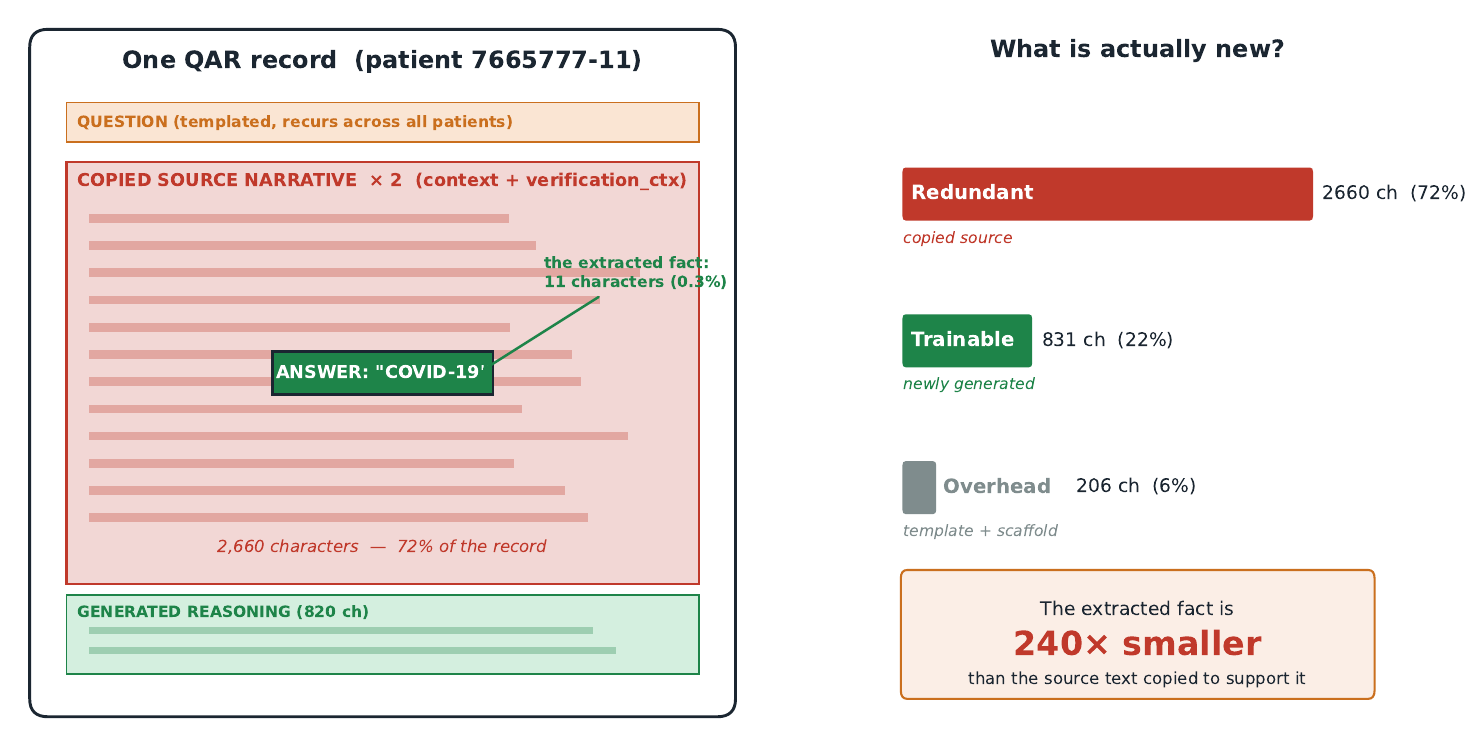}
\caption{\textbf{Context-copy redundancy in a single real record.} One
question-answering-with-reasoning (QAR) item from the corpus (patient
\texttt{7665777-11}), shown by its constituent field sizes. The extracted
clinical fact, the eleven-character answer ``COVID-19'', is stored alongside
2{,}660 characters of verbatim source narrative (the \texttt{context} field and
a second copy in the verification metadata). Of the record's 3{,}697 characters,
72\% are copied source and only 22\% are newly generated, of which the extracted
fact itself is just 0.3\%; the single extracted fact is roughly 240 times smaller
than the source text copied to support it. This per-record pattern, multiplied
across the many items each patient generates, is the origin of the corpus-level
context-copy redundancy quantified in Table~\ref{tab:per-task-redundancy}.}
\label{fig:worked-example}
\end{figure}

This Supplementary Information provides additional results and full numerical
detail supporting the main manuscript, in the order presented below. It documents:
the complete experimental configuration (Supplementary
Table~\ref{tab:config}); a per-channel comparison of provenance redundancy
against compression-based redundancy (Supplementary
Table~\ref{tab:perchannel-supp}); a mechanism-level decomposition of redundancy
into copied context and duplicated generation for each channel (Supplementary
Table~\ref{tab:s2-mechanisms}); the field-to-category mapping used by the
provenance classifier (Supplementary Table~\ref{tab:fieldmap}); the
near-duplicate robustness of the exact-match redundancy estimate (Supplementary
Table~\ref{tab:neardup}); the downstream disease-NER results at the two shorter
adaptation depths that complement the primary results in the main text, for the
NCBI-Disease (Supplementary Table~\ref{tab:downstream-depths}) and BC5CDR-Disease
(Supplementary Table~\ref{tab:bc5cdr-depths}) benchmarks; and a worked example of
each redundancy mechanism drawn from a single patient record (Supplementary
Table~\ref{tab:mechanisms}). The accompanying figures present the mechanism-level
decomposition (Supplementary Figure~\ref{fig:s1}), the masked-language-model
adaptation-loss trajectories (Supplementary Figure~\ref{fig:loss}), and a
single-record illustration of context-copy redundancy (Supplementary
Figure~\ref{fig:worked-example}). All values are reported consistently with the
main text; compression ratios are computed on the full corpus, and for the
dictionary and block-sorting compressors they agree with subsample-based
estimates to within an absolute difference of $0.001$ in the compression ratio.
Together, these analyses support the main finding that the
majority of the corpus consists of non-trainable redundancy arising through two
distinct mechanisms, copied context and duplicated generation.

\subsection{Corpus construction and availability}

The corpus analysed here is drawn from the output of a multi-agent clinical
extraction pipeline applied to the 167{,}034 patient narratives of PMC-Patients.
The pipeline pairs two large language models: a generator (Llama-3.3-70B) that
produces structured clinical artifacts from each narrative, and a clinical-domain
verifier (MMed-Llama-3.1-70B) that checks each generated artifact against the
source text before it is retained. Generation is deliberately grounded in the
source narrative so that every artifact remains traceable to, and verifiable
against, the originating text; reproducing source spans within items is a
designed property of this grounding, not an incidental one. The pipeline
produces a range of outputs per patient, including a full structured report and
a set of task-specific artifacts from which further downstream tasks can be
derived. The present analysis concerns the ten text-bearing extraction channels
that carry generated clinical content, namely named-entity recognition, relation
extraction, question answering with reasoning, temporal events, summarisation,
medications, risk question answering, recommendations, risk states, and risks; a
non-textual visualisation payload is excluded. Model and tokenisation settings
are given in Supplementary Table~\ref{tab:config}, and the mapping from JSON
fields to provenance categories in Supplementary Table~\ref{tab:fieldmap}. The
pipeline's design and capabilities are beyond the scope of this paper, which
characterises the provenance structure of its extraction output rather than the
system that produced it.

The grounding-and-verification design is what gives the corpus its provenance
structure, and the redundancy this paper measures is an inherent and expected
consequence of faithful grounded extraction rather than a deficiency of the
output. Because each artifact is anchored to the source narrative for
verifiability, the corpus necessarily reproduces source text (copied context);
because the same clinical facts about a patient legitimately recur across
multiple extraction views, generated content is necessarily repeated (duplicated
generation). Both are structural consequences of grounding and multi-view
extraction that any faithful pipeline would exhibit, and the grounding they
reflect is what supports verifiable, hallucination-resistant extraction. Our
contribution is to \emph{measure} this structure, so that the size of a grounded
clinical corpus is not mistaken for the quantity of unique information it
carries. The corpus is a research artifact produced for this study and is not
yet publicly released; it can be regenerated from the publicly available
PMC-Patients narratives using the released analysis code, and will be made
available subject to the data terms of the source corpus.


\begin{table}[htbp]
\centering
\caption{Complete experimental configuration. All settings were fixed before
analysis and are reported here so the measurement and downstream experiment can
be reproduced exactly. Token counts use the Llama-3.3-70B tokenizer throughout.}
\label{tab:config}
\footnotesize
\renewcommand{\arraystretch}{1.15}
\begin{tabularx}{\textwidth}{@{} l X @{}}
\toprule
\textbf{Component} & \textbf{Setting} \\
\midrule
\multicolumn{2}{@{}l}{\textbf{\textit{Pipeline and corpus}}} \\
Source corpus              & PMC-Patients (167{,}034 patient narratives) \\
Generator model            & Llama-3.3-70B \\
Verifier model             & MMed-Llama-3.1-70B \\
Text-bearing channels      & 10 (of 11; visualisation-payload channel excluded) \\
Tokenizer                  & Llama-3.3-70B \\
\addlinespace[0.3em]
\multicolumn{2}{@{}l}{\textbf{\textit{Provenance decomposition (PRD)}}} \\
Classification unit        & Per string-valued field via recursive JSON walk \\
Source de-duplication      & Exact-match MD5, corpus-wide, counted once \\
Generated de-duplication   & Exact-match MD5, corpus-wide (first occurrence unique) \\
Copied context             & Not de-duplicated (every copy counted) \\
\addlinespace[0.3em]
\multicolumn{2}{@{}l}{\textbf{\textit{Compression corroboration}}} \\
Compressors                & gzip (level 6), bzip2 (level 9), LZMA (preset 6) \\
PPMD                       & Model order 16, 2048\,MB memory \\
Scope                      & Full corpus (all four provenance streams) \\
Near-duplicate check       & MinHash, Jaccard $>0.85$ \\
\addlinespace[0.3em]
\multicolumn{2}{@{}l}{\textbf{\textit{Encoder adaptation (continued MLM)}}} \\
Backbone                   & BioClinical ModernBERT-base (150\,M parameters) \\
Masking probability        & 0.15 \\
Batch size                 & 32 \\
Maximum sequence length    & 256 \\
Learning rate              & $5\times10^{-5}$ (200 warmup steps) \\
Adaptation depths          & 10{,}000 / 20{,}000 / 40{,}000 steps (40k primary) \\
Token budget per condition & 174.3\,M whitespace tokens (equal-budget control) \\
Seeds                      & 3 per condition per depth \\
Precision                  & Mixed precision \\
\addlinespace[0.3em]
\multicolumn{2}{@{}l}{\textbf{\textit{Linear probe (downstream evaluation)}}} \\
Probe                      & Frozen backbone, single linear head \\
Epochs                     & 5 \\
Learning rate              & $10^{-3}$ \\
Maximum sequence length    & 256 \\
Benchmarks                 & NCBI-Disease, BC5CDR-Disease (disease-only splits) \\
Metric                     & seqeval entity F1 \\
Statistics                 & Mean $\pm$ s.d.\ over 3 seeds; 1{,}000-resample bootstrap CIs (2{,}000 for overall-gain significance) \\
\addlinespace[0.3em]
\multicolumn{2}{@{}l}{\textbf{\textit{Hardware}}} \\
GPU                        & NVIDIA 4 $\times$ H200 (141\,GB each) \\
Scheduler                  & SLURM (Supercomputing Wales Falcon) \\
\bottomrule
\end{tabularx}
\end{table}

\begin{table}[htbp]
\centering
\caption{Per-channel comparison of provenance redundancy (copied context plus
duplicated generation, \% of channel tokens) against compression reduction
(\%, LZMA). The two measures agree on the
copied-context-rich channels (QAR, RE) and diverge where text is
stylistically predictable but not copied (e.g.\ the summary channel). The
source-narrative channel is excluded, as it is the reference text against
which redundancy is defined rather than a generated channel.}
\label{tab:perchannel-supp}
\renewcommand{\arraystretch}{1.2}
\begin{tabularx}{\textwidth}{@{} X R R @{}}
\toprule
\textbf{Channel} & \textbf{Provenance redundancy (\%)} & \textbf{Compression reduction (\%)} \\
\midrule
QAR (question answering with reasoning) & 90.7 & 96.9 \\
RE (relation extraction)                & 87.0 & 96.3 \\
Recommendations                         & 70.6 & 97.9 \\
Risks                                   & 68.1 & 87.1 \\
Temporal events                         & 68.0 & 80.6 \\
Risk states                             & 66.1 & 92.0 \\
Risk-QA                                 & 51.3 & 95.5 \\
Medications                             & 25.9 & 79.4 \\
NER (named-entity recognition)          & 19.8 & 75.7 \\
Summary                                 &  0.0 & 81.0 \\
\bottomrule
\end{tabularx}
\end{table}

\begin{table}[htbp]
\centering
\caption{Mechanism-level decomposition of provenance redundancy by channel.
Values are percentages of channel tokens attributed to copied context and to
duplicated generation; their sum is the total redundancy. Copied context
dominates the QAR and RE channels, whereas several downstream channels
(e.g.\ recommendations, risks, risk-QA) are dominated by duplicated
generation. All values are from the full 167{,}034-patient provenance
analysis.}
\label{tab:s2-mechanisms}
\renewcommand{\arraystretch}{1.2}
\begin{tabularx}{\textwidth}{@{} X R R R @{}}
\toprule
\textbf{Channel} & \textbf{Copied context (\%)} & \textbf{Duplicated generation (\%)} & \textbf{Total redundancy (\%)} \\
\midrule
QAR (question answering with reasoning) & 81.0 &  9.7 & 90.7 \\
RE (relation extraction)                & 80.0 &  7.0 & 87.0 \\
Recommendations                         &  0.0 & 70.6 & 70.6 \\
Risks                                   &  0.0 & 68.1 & 68.1 \\
Temporal events                         & 47.1 & 20.9 & 68.0 \\
Risk states                             & 20.1 & 46.0 & 66.1 \\
Risk-QA                                 &  0.0 & 51.3 & 51.3 \\
Medications                             &  0.0 & 25.9 & 25.9 \\
NER (named-entity recognition)          &  0.0 & 19.8 & 19.8 \\
Summary                                 &  0.0 &  0.0 &  0.0 \\
\bottomrule
\end{tabularx}
\end{table}

\begin{table}[htbp]
\centering
\caption{Field-to-category mapping used by Provenance-based Redundancy
Decomposition. Every string-valued field in the pipeline output is assigned to
one of five provenance categories by the rules below; the assignment was fixed
in advance from the output schema and applied uniformly to all records and all
ten text-bearing channels. The same taxonomy drives both the token classifier
and the compression serialisation, so the two analyses cover identical text.}
\label{tab:fieldmap}
\renewcommand{\arraystretch}{1.25}
\begin{tabularx}{\textwidth}{@{} l X @{}}
\toprule
\textbf{Category} & \textbf{Fields / rule} \\
\midrule
Unique source &
The top-level \texttt{text} field (patient narrative); de-duplicated corpus-wide
by MD5 and counted once, regardless of how many records contain it. \\
\addlinespace
Copied context &
\texttt{context}, \texttt{verification\_anchor}, \texttt{verification\_ctx},
\texttt{sentence}. Reproduce the source narrative verbatim and are counted as
redundant by construction (not de-duplicated; every copy counts). A placeholder
value (\texttt{Context unavailable}, empty, \texttt{n/a}, \texttt{none},
\texttt{.}) is reassigned to scaffold. \\
\addlinespace
Scaffold &
Identifiers and enumerated metadata:
\texttt{qa\_id}, \texttt{rel\_id}, \texttt{event\_id}, \texttt{uid},
\texttt{start\_char}, \texttt{end\_char}, \texttt{confidence},
\texttt{llm\_confidence}, \texttt{direction}, \texttt{temporal\_order},
\texttt{is\_anchored}, \texttt{is\_negated\_source}, \texttt{timepoint\_type},
\texttt{event\_type}, \texttt{assertion\_status}, \texttt{verifier\_status},
\texttt{status}, \texttt{head\_type}, \texttt{tail\_type}, \texttt{label},
\texttt{category}, \texttt{meta\_species}, \texttt{verdict\_path},
\texttt{med\_id}, \texttt{drug}, \texttt{route}, \texttt{frequency},
\texttt{qtype}, \texttt{risk\_id}, \texttt{state}, \texttt{severity},
\texttt{volatility\_profile}, \texttt{rule}, \texttt{threshold},
\texttt{decision\_type}, \texttt{severity\_marker}, \texttt{method},
\texttt{actionability}, \texttt{source}, \texttt{provenance},
\texttt{last\_updated}, \texttt{rec\_id}, \texttt{type},
\texttt{agentic\_check}, \texttt{constraints\_applied},
\texttt{based\_on\_risks}, \texttt{level}, \texttt{node\_type}, \texttt{size}. \\
\addlinespace
Unique / duplicated generated &
All remaining string fields (answers, reasoning chains, summaries, entity
strings). A single corpus-wide MD5 set spans all ten channels: the first
occurrence of a string anywhere in the corpus is unique generated content, and
every subsequent identical occurrence, including across different channels, is
duplicated generated content. \\
\bottomrule
\end{tabularx}
\end{table}

\begin{table}[htbp]
\centering
\caption{Near-duplicate robustness of the exact-match redundancy estimate.
Applying MinHash near-duplicate detection (Jaccard $>0.85$) to a 2\% sample of
generated content (76{,}400 strings) reduces the unique fraction only
marginally relative to exact matching, and relaxing the match to number-only
differences accounts for a negligible share. The exact-match figures reported
throughout are therefore a conservative lower bound on redundancy, and the
generated-content redundancy is predominantly verbatim rather than paraphrastic.}
\label{tab:neardup}
\renewcommand{\arraystretch}{1.2}
\begin{tabularx}{\textwidth}{@{} X R @{}}
\toprule
\textbf{Matching criterion} & \textbf{Generated strings marked unique (\%)} \\
\midrule
Exact match (used throughout)            & 57.8 \\
MinHash near-duplicate (Jaccard $>0.85$) & 56.3 \\
\midrule
\textit{Difference (near-dup vs exact)}  & \textit{1.5 pts} \\
\textit{\quad of which number-only differences} & \textit{0.2 pts} \\
\bottomrule
\end{tabularx}
\end{table}

\begin{table}[htbp]
\centering
\caption{Downstream disease-NER F1 (NCBI-Disease test; mean $\pm$ s.d.\ over
three seeds) at the two shorter adaptation depths used as a robustness ladder,
complementing the 40{,}000-step primary results in the main text
(Table~\ref{tab:downstream}). The overall and common-disease de-duplication
gains are stable across depths; the rare-slice gain is small and changes
substantially across depths (from $+0.035$ at 10{,}000 steps to $+0.001$ at
20{,}000), and the difference-in-differences between the rare and common slices
is negative at 20{,}000 steps, consistent with the absence of a reliable
rare-specific effect reported in the main text.}
\label{tab:downstream-depths}
\footnotesize
\renewcommand{\arraystretch}{1.2}
\begin{tabularx}{\textwidth}{@{} l l C C C C @{}}
\toprule
\textbf{Depth} & \textbf{Condition} & \textbf{Rare} & \textbf{Common} & \textbf{Unseen} & \textbf{All} \\
 & & ($n{=}152$) & ($n{=}441$) & ($n{=}367$) & ($n{=}960$) \\
\midrule
\multirow{3}{*}{10{,}000}
 & Raw (redundant)        & $0.805 \pm 0.018$ & $0.835 \pm 0.019$ & $0.432 \pm 0.024$ & $0.665 \pm 0.015$ \\
 & De-duplicated          & $0.840 \pm 0.017$ & $0.863 \pm 0.005$ & $0.456 \pm 0.019$ & $0.692 \pm 0.009$ \\
 & Context-removed (abl.) & $0.820 \pm 0.029$ & $0.855 \pm 0.010$ & $0.449 \pm 0.037$ & $0.684 \pm 0.024$ \\
\cmidrule(l){2-6}
 & \textit{De-dup.\ gain} & $+0.035$ & $+0.028$ & $+0.023$ & $+0.027$ \\
\midrule
\multirow{3}{*}{20{,}000}
 & Raw (redundant)        & $0.818 \pm 0.015$ & $0.846 \pm 0.012$ & $0.444 \pm 0.021$ & $0.677 \pm 0.015$ \\
 & De-duplicated          & $0.818 \pm 0.015$ & $0.873 \pm 0.016$ & $0.455 \pm 0.027$ & $0.694 \pm 0.017$ \\
 & Context-removed (abl.) & $0.824 \pm 0.034$ & $0.864 \pm 0.015$ & $0.443 \pm 0.028$ & $0.686 \pm 0.022$ \\
\cmidrule(l){2-6}
 & \textit{De-dup.\ gain} & $+0.001$ & $+0.028$ & $+0.011$ & $+0.017$ \\
\bottomrule
\end{tabularx}
\end{table}

\begin{table}[htbp]
\centering
\caption{Replication on BC5CDR-Disease: downstream disease-NER F1
(mean $\pm$ s.d.\ over three seeds) for encoders adapted on the raw,
de-duplicated, and context-removed corpus at equal token budget, at the two
shorter adaptation depths used as a robustness ladder; the 40{,}000-step primary
results are in the main text (Table~\ref{tab:downstream}). As on NCBI-Disease,
de-duplication improves the encoder overall and on common diseases across depths.
Mentions are stratified by corpus disease frequency using the same
bottom-quartile rule as for NCBI; the dataset-relative threshold is corpus
frequency $\leq 30$. The corpus covered $36.1\%$ of BC5CDR's distinct test
diseases, lower than its NCBI coverage, reflecting BC5CDR's broader disease
range; absolute F1 is therefore not comparable across the two benchmarks, but
the de-duplication \emph{gain} is.}
\label{tab:bc5cdr-depths}
\footnotesize
\renewcommand{\arraystretch}{1.2}
\begin{tabularx}{\textwidth}{@{} l l C C C C @{}}
\toprule
\textbf{Depth} & \textbf{Condition} & \textbf{Rare} & \textbf{Common} & \textbf{Unseen} & \textbf{All} \\
 & & ($n{=}353$) & ($n{=}2016$) & ($n{=}2055$) & ($n{=}4424$) \\
\midrule
\multirow{3}{*}{10{,}000}
 & Raw (redundant)        & $0.590 \pm 0.015$ & $0.822 \pm 0.016$ & $0.619 \pm 0.021$ & $0.600 \pm 0.015$ \\
 & De-duplicated          & $0.606 \pm 0.009$ & $0.836 \pm 0.012$ & $0.624 \pm 0.009$ & $0.617 \pm 0.005$ \\
 & Context-removed (abl.) & $0.591 \pm 0.008$ & $0.820 \pm 0.014$ & $0.616 \pm 0.018$ & $0.603 \pm 0.004$ \\
\cmidrule(l){2-6}
 & \textit{De-dup.\ gain} & $+0.016$ & $+0.014$ & $+0.005$ & $+0.016$ \\
\midrule
\multirow{3}{*}{20{,}000}
 & Raw (redundant)        & $0.579 \pm 0.018$ & $0.824 \pm 0.018$ & $0.607 \pm 0.007$ & $0.603 \pm 0.013$ \\
 & De-duplicated          & $0.599 \pm 0.009$ & $0.834 \pm 0.012$ & $0.624 \pm 0.009$ & $0.616 \pm 0.004$ \\
 & Context-removed (abl.) & $0.615 \pm 0.004$ & $0.822 \pm 0.015$ & $0.627 \pm 0.011$ & $0.611 \pm 0.004$ \\
\cmidrule(l){2-6}
 & \textit{De-dup.\ gain} & $+0.020$ & $+0.011$ & $+0.017$ & $+0.013$ \\
\bottomrule
\end{tabularx}
\end{table}

\begin{table}[htbp]
\centering
\caption{The two redundancy mechanisms, with examples drawn from one
patient record (uid 7665777-11, the record used in
Figure~\ref{fig:schematic}) and corpus-wide counts. Context-copy
redundancy reproduces the same source span verbatim across many of a
patient's items; generation-duplication redundancy re-emits near-identical
model output across records. Only the first is removable by restructuring
storage.}
\label{tab:mechanisms}
\footnotesize
\renewcommand{\arraystretch}{1.3}
\begin{tabularx}{\textwidth}{@{} >{\bfseries}l X X @{}}
\toprule
 & \textbf{Context-copy redundancy} & \textbf{Generation-duplication redundancy} \\
\midrule
What recurs & A source-narrative span copied into a field & Templated model output re-emitted across records \\
Field(s) & \texttt{context}, \texttt{sentence} & \texttt{answer}, \texttt{reason}, entity strings \\
Example &
The narrative span ``\emph{This 77-year-old male patient was transferred
to our ICU one week after his COVID-19 diagnosis due to continuing
respiratory decompensation requiring intubation\ldots}'' is copied
verbatim into the \texttt{context} field of six separate items for this
patient &
The same generated reasoning scaffold recurs corpus-wide: the
extraction-rule wrapper (\texttt{LLM\_Extraction} / \texttt{DIRECT\_EXTRACTION})
appears in 1{,}035{,}199 items, and the reasoning opener
``\emph{\ldots is significant as it\ldots}'' in 64{,}903, with only the
clinical entity varied \\
Removable? & Yes: store each source span once and reference it & No: intrinsic to how the model writes \\
\bottomrule
\end{tabularx}
\end{table}

\subsection{Notes on Interpretation}
Provenance redundancy and compression quantify related but distinct
properties of the data. Provenance identifies the origin of each token
(copied, duplicated, or novel), whereas compression measures statistical
predictability regardless of origin. Channels dominated by copied context
(QAR, RE) show strong agreement between the two measures. Channels with low
provenance redundancy may nonetheless compress well, because clinical text
reuses shared phrasing and structure across records even when each item is a
distinct generation; the summary channel, which carries no copied context yet
compresses by 81\%, is the clearest example. This pattern reflects
complementary perspectives on redundancy rather than disagreement between the
methods.

The mechanism-level values in Supplementary Table~\ref{tab:s2-mechanisms}
show that copied-context redundancy dominates the QAR and RE channels, whereas
duplicated generation is the primary source of redundancy in several
downstream channels. The summary channel exhibits neither mechanism,
consistent with its role as an internal control in the main text.


\end{appendices}

\end{document}